\pdfoutput=1

\documentclass{article}

\PassOptionsToPackage{numbers, compress}{natbib}

\usepackage[final]{neurips_2023}

\usepackage[utf8]{inputenc} 
\usepackage[T1]{fontenc}    
\usepackage{hyperref}       
\usepackage{url}            
\usepackage{booktabs}       
\usepackage{amsfonts}       
\usepackage{nicefrac}       
\usepackage{microtype}      
\usepackage{xcolor}         
\usepackage{xspace}
\usepackage{amssymb}
\usepackage[export]{adjustbox}
\usepackage{multirow,booktabs,makecell}
\usepackage{floatrow}
\usepackage{mathabx}
\usepackage{enumitem}
\usepackage{xcolor}
\usepackage{adjustbox}
\usepackage{caption}
\usepackage{subcaption}
\usepackage{array}
\usepackage{comment}

\newcommand{\method}{\textsc{LSKD}\xspace}
\newcommand{\eg}{\textit{e}.\textit{g}.\xspace}
\newfloatcommand{capbtabbox}{table}[][\FBwidth]

\title{Localized Symbolic Knowledge Distillation \\for Visual Commonsense Models}
\author{
\vspace{1mm}
\textbf{Jae Sung Park}\textsuperscript{1},
\textbf{Jack Hessel}\textsuperscript{2},
\textbf{Khyathi Raghavi Chandu}\textsuperscript{2}, \\
\vspace{1mm}
\textbf{Paul Pu Liang}\textsuperscript{2,4},
\textbf{Ximing Lu}\textsuperscript{1,2},
\textbf{Peter West}\textsuperscript{1,2}, \\
\textbf{Youngjae Yu}\textsuperscript{5},
\textbf{Qiuyuan Huang}\textsuperscript{3},
\textbf{Jianfeng Gao}\textsuperscript{3},
\textbf{Ali Farhadi}\textsuperscript{1,2},
\textbf{Yejin Choi}\textsuperscript{1,2}
\\\\ 
\textsuperscript{1}University of Washington
\textsuperscript{2}Allen Institute for Artificial Intelligence
\textsuperscript{3}Microsoft Research\\
\textsuperscript{4}Carnegie Mellon University
\textsuperscript{5}Yonsei University
}

\begin{document}
\maketitle{}
\vspace{-0.5cm}

\begin{abstract}

Instruction following vision-language (VL) models offer a flexible interface that supports a broad range of multimodal tasks in a zero-shot fashion.
However, interfaces that operate on full images do not directly enable the user to ``point to" and access specific regions within images. This capability is important not only to support reference-grounded VL benchmarks, but also, for practical applications that require precise \emph{within-image} reasoning. We build Localized Visual Commonsense models, which allow users to specify (multiple) regions as input. We train our model by sampling localized commonsense knowledge from a large language model (LLM): specifically, we prompt an LLM to collect commonsense knowledge given a \emph{global} literal image description and a \emph{local} literal region description automatically generated by a set of VL models. 
With a separately trained critic model that selects high-quality examples, we find that training on the localized commonsense corpus can successfully distill existing VL models to support a reference-as-input interface.  Empirical results and human evaluations in a zero-shot setup demonstrate that our distillation method results in more precise VL models of reasoning compared to a baseline of passing a generated referring expression to an LLM \footnote{Code will be released in \url{https://github.com/jamespark3922/localized-skd}}. 
\end{abstract}
\section{Introduction}

Large language models are capable of efficiently performing a wide array of tasks in a zero-shot fashion. 
For text-only models, one commonly adopted interface is a flexible, language specification of inputs coupled with an imperative request, \eg, ``\texttt{[article text]. Summarize this article.}"
Similarly, a natural extension allowing visual inputs manifests as, \eg, ``\texttt{[image]. Describe this image}".

However, as models expand beyond text-only modalities, they should incorporate more flexible forms of user input as well. Allowing users to specify individual objects/actors/regions within an image as part of the input query is an important challenge, e.g., the \texttt{[image] [request]} interface above would not directly a user to ask \texttt{Why is [this person in the image] sad?}.
One option would be to simply require users specifically describe the piece of the image they are attempting to specify, \eg, ``\texttt{[image] [description of specific region] [request]}". However, authoring concrete referring expressions is not only cumbersome, particularly for scenes with lots of objects (e.g., ``the person in the red jacket on the left of the scene with their arms crossed") but also challenging, even for humans: \cite{dale1995computational} argue that a good referring expression should both specify the reference precisely, but also, follow Grice's maxim of Quantity, i.e., provide no extra information. Given this tension, many popular referring expression datasets are gathered in a sophisticated ``gamified" fashion \cite{von2006peekaboom,kazemzadeh2014referitgame}, which aims to balance underspecification vs. verbosity.

We argue instead that users of vision-augmented LLMs should instead be able to pass localized visual references simply by ``pointing" to regions within the image \cite{bolt1980put,siroux1995modeling,oviatt2007multimodal}.
This enables models to focus on the region while interpreting the user's request in a more intuitive fashion, and provide more accurate and contextually relevant responses. By incorporating localized visual references, the model can better understand and interpret complex scenes, thereby improving its performance on tasks requiring a detailed understanding of the visual context.



We propose Localized Symbolic Knowledge Distillation (\method): the core idea is to provide literal descriptions of images to a large language model and allow that model to connect the dots between these literal descriptors (e.g., lists of objects) and a holistic perspective of the scene. Different from recent works which also distill from an LLM conditioned on visual descriptors symbolically \cite{liu2023visual,zhu2023minigpt}, we additionally provide a localized reference to a particular region within the image and design prompts to encourage the LLM to generate commonsense inference about that specific region. After sampling, we train Localized Visual Commonsense models to generate commonsense triples conditioned on the image and the region directly; we show that this process effectively distills the LLM's capacity for global+local scene understanding highlighted by zero-shot results on localized visual reasoning benchmarks and human evaluation.

In summary, our main contributions are:
\begin{enumerate}[leftmargin=*,topsep=0pt,itemsep=-1ex,partopsep=1ex,parsep=1ex]
    \item A new scalable framework that can generate reliable and localized visual commonsense statements. 
    
    \item \textit{The Localized Commonsense Knowledge Corpus}: 1M localized commonsense inferences posed over 250K images. This dataset can be used to expand the capacity of existing vision+language models to incorporate references-as-input with no architectural modifications.
    
    \item Achieving the SoTA zero-shot performance for three localized visual reasoning tasks.
    
    \item Human evaluation results suggesting that a strong student model outperforms the teacher model in answering localized visual commonsense questions.

\end{enumerate}


% ...

\section{Distilling Localized Visual Commonsense from a LLM}
Here, we describe our LSKD pipeline to distill visual commonsense from a LLM. Prior works have explored powerful LLM as the teacher model (GPT-3, ChatGPT) to apply knowledge distillation for language-only reasoning tasks \cite{west2021symbolic,DBLP:conf/emnlp/LiuSSC22, Bhagavatula2022I2D2IK}. Multimodal inputs offer additional challenges in grounding regions to relevant texts. Our work addresses this challenge by automatically generating reliable and diverse knowledge statements for multimodal input, to further reason about regions within an image.

\begin{figure*}[h]
\centering
\includegraphics[trim=2cm 1cm 0cm 0cm, width=0.99\linewidth]{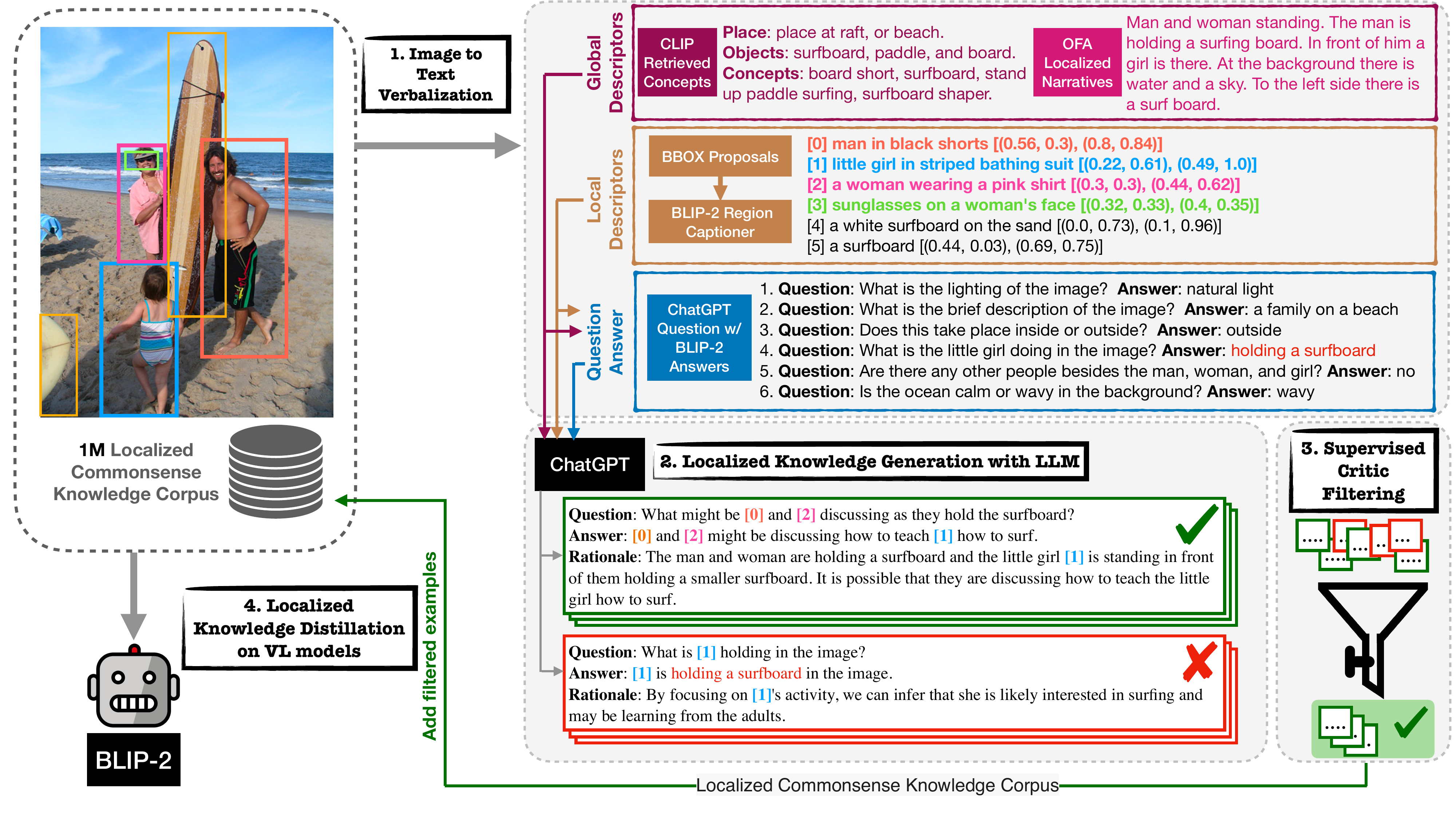}
\caption{Pipeline of our \method framework. 1) Diverse vision-language descriptors are used to verbalize images. 2) LLMs leverage the global and local descriptors to generate grounded commonsene knowledge. 3) We annotate a small subset of data to train a supervised critic model that can filter instances displaying incorrect visual details or incoherent reasoning. The critic model filters the rest of generated statements to finalize the data pool. 4) A multimodal model is finetuned on the synthetic data to support localized visual commonsense reasoning in a zero-shot manner.
}
\label{fig:pipeline}
\end{figure*}

Figure~\ref{fig:pipeline} shows the overall framework of \method\footnote{For visualization purposes, we provide a shortened version of verbalizations. The full verbalization output with the prompt to call ChatGPT is shown in the Appendix.}. To learn from the LLM as our teacher model, we verbalize the image into a set of dense text statements generated by global descriptors that provide relevant, general overall semantics of the image, and local descriptors that talk about specific regions in the image. We then pass these automatically generated descriptions to LLM and prompt to mine localized, commonsense statements about the image at scale (See the Appendix for the exact prompt). 

As LLMs comprehend multimodal input only through machine-generated image-to-text verbalization, they are prone to hallucination and generation of inconsistent statements about the image. For instance, an incorrect verbalizer output, as in Figure~\ref{fig:pipeline}, might cause the LLM to produce visually incoherent statements like "[1] is holding a surfboard".
To minimize errors in modality translation, we construct a critic model, trained on a limited set of high-quality, hand-annotated instances to detect and remove such inconsistencies. This critic model mimics human judgment in evaluating the generated commonsense knowledge, so that we can intentionally oversample localized knowledge data, and utilize it to filter out non-relevant instances. Finally, we finetune a vision-language model on the high-quality synthetic data to facilitate zero-shot localized visual commonsense reasoning. We use 250K images in union of Visual Genome ~\cite{krishna2017visual} and VCR ~\cite{zellers2019recognition}, which include a diverse set of social situations involving people and objects, as the seed images to collect the knowledge corpus. After filtering, we collect 1M instances of Localized Commonsense Knowledge Corpus with information grounded to specific regions in the image (see Appendix \ref{sec:corpus_detail} for more details).


\subsection{Image to Text Verbalization}
We first describe our methods for verbalizing (i.e., writing out in natural language) images/regions to text. Note that this step does not require images with text annotations for the target datasets, unlike prior work ~\cite{liu2023visual}, and can be applied to any set of images. We focus on deriving \emph{global} image descriptions, \emph{local} region descriptions, and \emph{dynamic} question-answer pairs for each image. Figure~\ref{fig:pipeline} gives a schematic of our process which includes an example image verbalization.

\paragraph{Global descriptors: Image Verbalizers}

Following \cite{zeng2022socratic}, we use the CLIP-ViTL model in a zero-shot fashion to extract basic concept information about the image using a template. We retrieve places from the Place365 ~\cite{Zhou2016PlacesAI}, objects from TencentML-Images ~\cite{Wu2019TencentMA}, and concepts from OpenImages ~\cite{Kuznetsova2018TheOI} to arrive at global concepts. Specifically: we use the
top 2 places, the top 3 object labels, and the top 3 concepts. In addition to concepts, we also get narrative descriptions of the entire image. For this, we fine-tuned OFA-Huge \cite{wang2022ofa} on the Localized Narratives \cite{PontTuset_eccv2020} corpus, which pairs 849K images with multi-sentence descriptions (we ignore the mouse trace information and simply treat the task as image-to-text captioning). We sample 5 localized narratives for each image using a temperature of 1.1. 

\paragraph{Local descriptors: Region Verbalizers.}
Global descriptors alone often fail to capture the intricate details of specific regions within an image, leading to a potential bottleneck in understanding scenes with more visual precision and enabling localized reasoning. We employ local descriptors that provide more grounded visual statements. To do so, we sample bounding box regions for the image using region proposal models from object detection literature ~\cite{li2022exploring}. We then train a region captioning model that maps from (image, region) $\rightarrow$ description of the region. We fine-tuned the generative version of BLIP-2 \cite{li2023blip} with the FLAN-t5-xxl \cite{chung2022scaling} backbone. We trained on datasets that provide descriptions of regions within images. a combination of
RefCOCO/RefCOCO+/RefCOCOg \cite{yu2016modeling,mao2016generation}, Sherlock Clues-only \cite{hessel2022abduction} (277K), and VisualGenome \cite{krishna2017visual} (1.96M): all of these datasets provide descriptions of given regions within images. Following \cite{zellers2021merlot,yao2021cpt}, we render the bounding box in the image itself to allow the model access to the bounding box's location. More details of the local descriptors are in Appendix ~\ref{sec:local}.

\paragraph{Dynamic descriptors: Q+A Verbalizers}
Finally, to support a holistic understanding and enable models to dynamically probe for potentially missing context, we acquire more fine-grained details about the scene using a series of questions and answers. Following ~\cite{Zhu2023ChatGPTAB}, we prompt an LLM to ask short, simple questions conditioning on the global and local descriptors as context, and query BLIP-2 \cite{li2023blip} in a zero-shot fashion to answer the questions. We specifically collect 15 question/answer pairs for each image.

\subsection{Localized Commonsense Knowledge Generation}
\label{sec:sec_with_generation_process}

For all experiments, we use ChatGPT as our LLM,\footnote{\url{https://openai.com/blog/chatgpt}} though in principle, any instruction-tuned LLM could be used.
We use question-answering-rationale (QAR) for knowledge representations. QAR representations are flexible, and have been successfully adopted as meaning representations in areas ranging from formal semantics \cite{he2015question,michael2017crowdsourcing,klein2022qasem} to commonsense reasoning \cite{talmor2018commonsenseqa,zellers2019recognition}.

Given the verbalization of images, we prompt ChatGPT to come up with an interesting and complex question with the possible answer that requires rationale to justify the reasoning. We support two versions of localized knowledge generation. One that refers to specific regions in the image either by their assigned numerical IDs and bounding box coordinates (\eg \texttt{What is [2] doing in the image?}) for more precise localization, and one that uses descriptive phrases (\eg \texttt{What is [the woman wearing a pink shirt] doing in the image?}) for more contextual and detailed references. Qualitatively, we observe that the LLM is able to connect the IDs and the region descriptions successfully, and create a convincing set of localized commonsense knowledge corpus. For each image, we prompt ChatGPT three times to generate three unique QARs sequentially. We do this for ID-based and description-based references (see Appendix for the prompts), and collect 18 localized instances per image.




\begin{figure}
\begin{floatrow}

\raisebox{5ex}{
\capbtabbox
{
\normalsize
\centering
\scalebox{0.75}{
\begin{tabular}{c c | c c c}
\toprule
QA MSE & Rationale MSE & \textbf{Precision} & \textbf{Recall} & \textbf{F1}\\
\midrule
& &  64.7 & 64.2 & 64.3 \\ 
\checkmark & & 66.3 & 65.2 & 65.7 \\
& \checkmark & 66.0 & 64.3 & 64.8 \\
\checkmark & \checkmark & \textbf{66.8} & \textbf{65.7} & \textbf{66.0} \\
\bottomrule
\end{tabular}
}
}
{
    \caption{
        Analysis of BLIP-2 based critic model. We see that adding the multi-class regression loss further improves the performance of critic model. 
    } 
    \label{tab:critic}
}

}

\ffigbox[\FBwidth]
  {\includegraphics[width=0.45\textwidth]{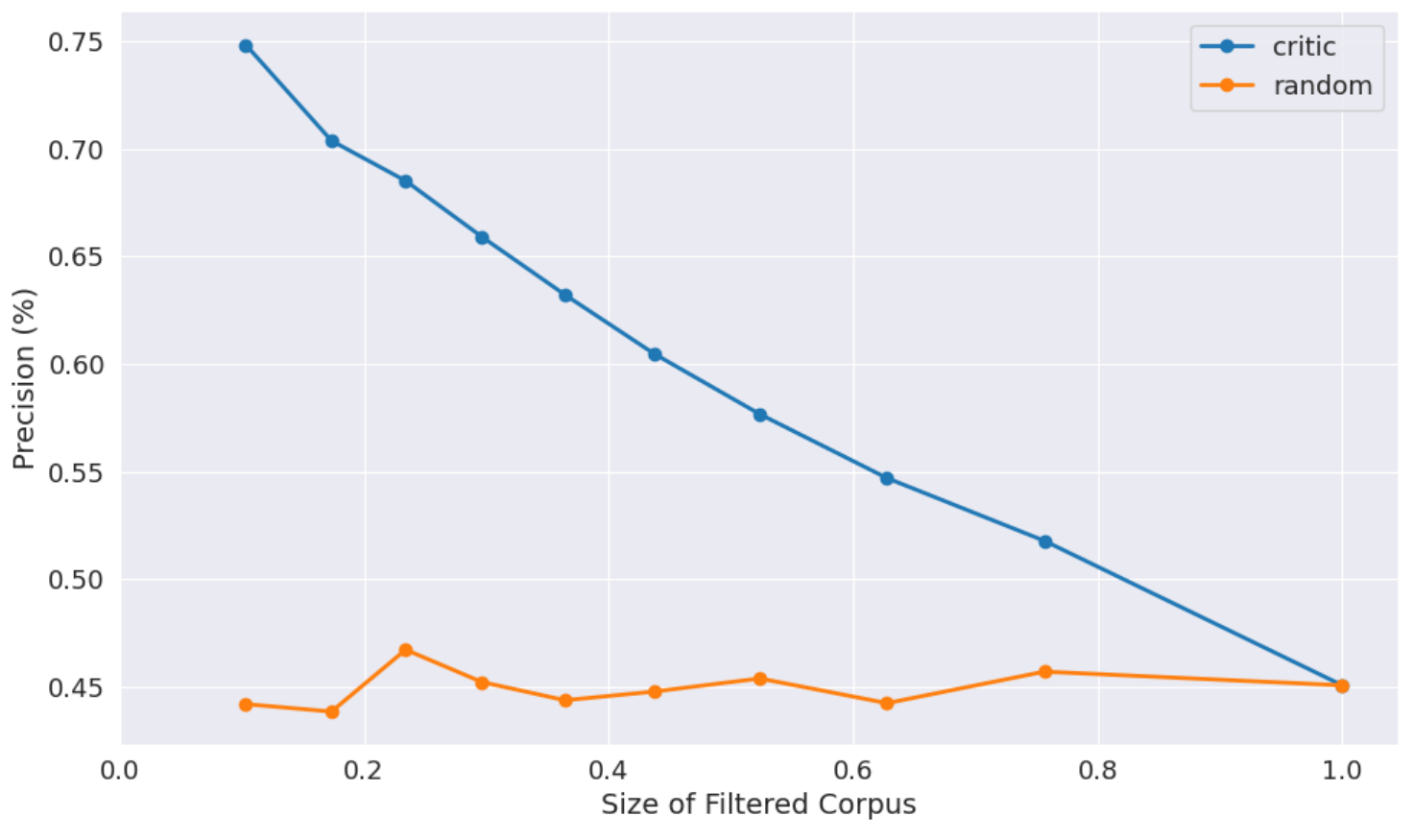}}
  {\caption{Precision of Critic Model with varying threshold values to filter the corpus size. Precision is increased significantly by using the supervised critic model to filter the corpus.}
  \label{fig:critic_filter}
  }
  
\end{floatrow}
\end{figure}



\subsection{Training the Critic Model}
\label{sec:critic}
 
 We train a supervised critic model to reflect the human acceptability of generated data. We allocate a subset of ~20K statements to train the critic model, and ~4k for evaluation. The ``accepted" instances should generally deliver the visually correct information and exhibit coherent reasoning. For each QAR, we ask two human annotators to rate from 1 to 3 (reject / maybe / accept) if 1) the QA displays visually correct information (QA rating), and 2) the rationale justifies the answer while being aligned with the image (QA $\to$ R rating)\footnote{The second criterion is automatically rejected if the QA has already rejected in the first pass}. We then assign binary label if at least one annotator has included reject for any of the two rating criteria. Using this labeling scheme, we found that only 45$\%$ of the instances are labeled as accepted, suggesting that aggressive filtering by the critic model is required.

For the model, we use a stage-1 pre-trained BLIP2 ~\cite{li2023blip} with ViT-G ~\cite{Fang2022EVAET} image encoder to do the critique. Following their finetuning scheme on retrieval tasks, we train the image encoder and Q-Former together, not freezing the weights. We add a linear layer to the image-text matching head that has been pre-trained to capture the multimodal content, and train it to perform the classification. 

We utilize the two rating criteria (QA and QA $\to$ R) to further inform the critic model to know what caused the humans to reject the QARs. We achieve this by multi-task training of critic model. The ratings containing reject are given the regression label of 0, while the average of two QA and QA $\to$ R ratings is calculated to get the regression label $y_{QA}$ and $y_{QA\to R}$. Along with the binary classification loss, the image-text matching head is further trained with mean squared error (MSE) losses with $y_{QA}$ and $y_{QA\to R}$. Table ~\ref{tab:critic} shows the performance of critic model on the above train and eval split. We empirically see that adding the multi-task loss (QS MSE and Rationale MSE) further helps the performance of classification.

\paragraph{Analysis of Supervised Critic} 
How reliable is the critic model on filtering erroneous instances? In the annotation stage, we have observed that only 45\% of the instances would be considered as valid by humans. We explore tuning different thresholds of critic model to filter the data (\eg keep instances whose predicted scores are higher than the threshold), and see if higher acceptability can be achieved with higher threshold. Figure~\ref{fig:critic_filter} shows a plot of precision value (instances labeled as ``accept") by the filtered corpus size. We see a consistent trend where removing the corpus with more critical criteria yields higher acceptability. Specifically, it jumps from 45\% of 70\% acceptance if 20\% are maintained by the critic model. We use this threshold value of 0.8 to apply the critic model. Note that filtering the corpus randomly, on the other hand, doesn't have any influence on the acceptability. 

\begin{figure}
\begin{floatrow}

\raisebox{5ex}{
\capbtabbox
    {
        \normalsize
        \centering
        \scalebox{0.75}{
            \begin{tabular}{|l|c|}
            \hline
            \textbf{Descriptors Used} & \textbf{Average Critic Score} \\ \hline
            Full Descriptors & 58.4 \\ \hline
            (-) CLIP Concepts & 52.1 \\ 
            (-) Localized Narratives & 56.1 \\ 
            (-) Global Descriptors & 54.3 \\ \hline
            (-) Local Descriptors & 49.8 \\ \hline
            (-) QAs & 49.0 \\ \hline
            \end{tabular}            
        }
    }
    {
        \caption{Ablation study of the descriptors. We remove one of the descriptors from full descriptors when calling ChatGPT to generate the corpus, and calculate the average critic score to rate the generations (higher the better).} 
        \label{tab:descriptors}
    }

}

\ffigbox[\FBwidth]
  {\includegraphics[width=0.4\textwidth]{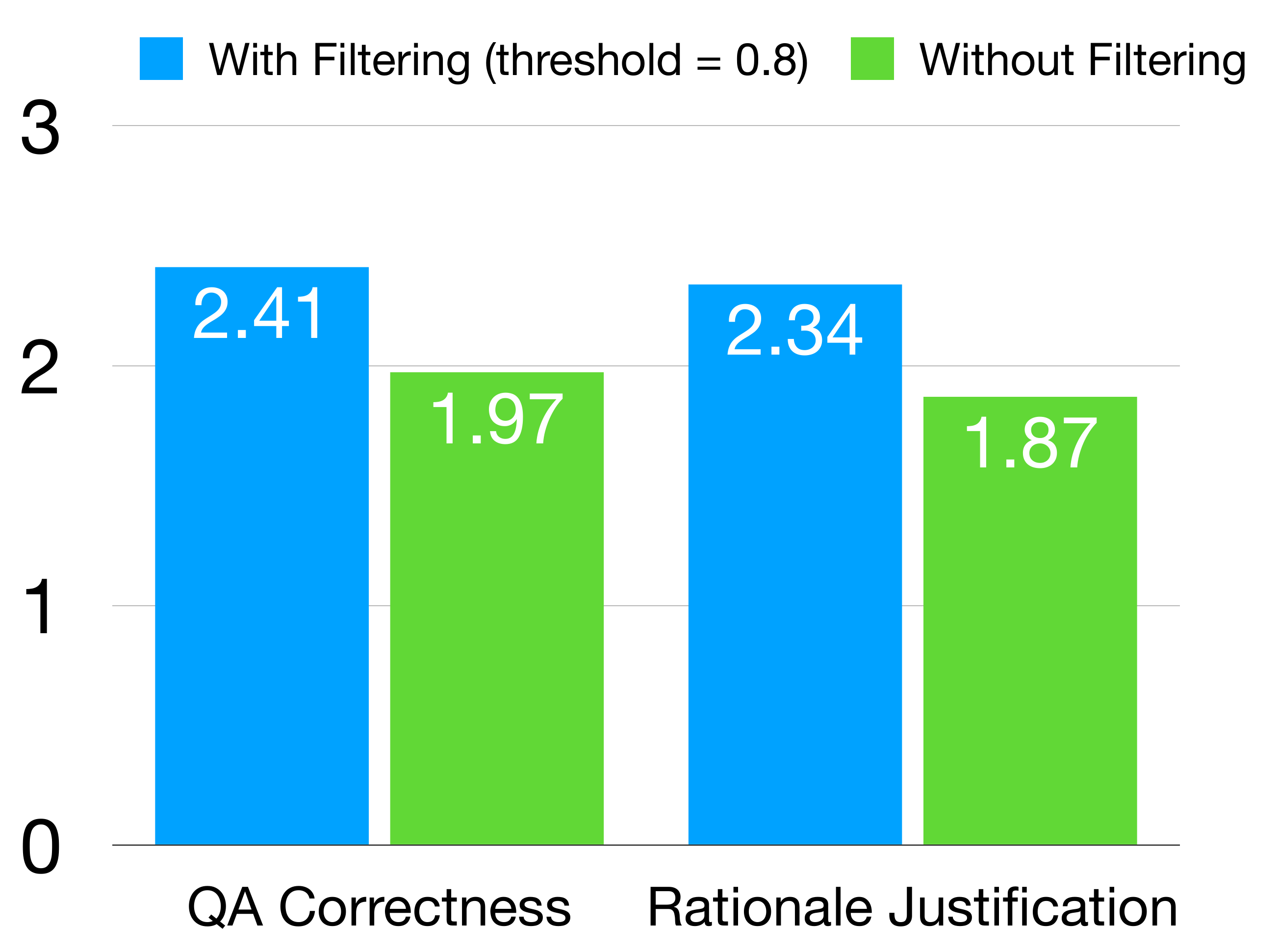}}
  {\caption{Human judgment of corpus with and without filtering. We get the average ratings in Likert scale (from 1 to 3) from three human annotators.}
  \label{fig:filtering_human_eval}
  }
  
\end{floatrow}
\end{figure}

In addition, we run human evaluation to measure the acceptability of data with and without filtering. We collect 500 instances the same way critic model labels are collected: 1) is the QA visually correct? and 2) does the rationale justify the answer? Likert scores from [1-3] are calculated for each criteria (higher the better). Figure ~\ref{fig:filtering_human_eval} shows the human evaluation results, and we see that the dataset with filtering is more favored by humans than without filtering.

\paragraph{Are All the Descriptors Necessary?}
We run ablation studies of the descriptor components in the ChatGPT prompt and use the critic model to score the ChatGPT generations. We collect QAR instances for 500 images and calculate the average critic score, with higher score aligned with human preference. Table~\ref{tab:descriptors} shows the result when one of the descriptors is removed from full verbalizations. We see that using all descriptors provides the best results, and in fact the QA descriptor provides the biggest jump (from 49.0 to 58.4).

\subsection{Training with the Localized Corpus}

We explore the distillation of localized commonsense knowledge by finetuning discriminative and generative vision language model on our corpus. For the corpus that mentions IDs and bounding box coordinates,  we follow ~\cite{zellers2021merlot, yao2021cpt, Zellers2022MERLOTRN, hessel2022abduction} by directly drawing colored highlights around the regions in the images where the region IDs and highlights are consistent throughout the corpus (\eg [0] always gets the color pink).  

During training, we additionally apply region-based augmentation by reassigning the IDs with a random order 
while still keeping a consistent color coding (\eg \textit{What might be [0] and [1] discussing?} $\to$ \textit{What might be [1] and [3] discussing?}). We similarly vary the number of regions to be shown in the image, while ensuring that the mentioned IDs are drawn in the image. With these tricks, the modifications are performed in the input image and text to enable localization, while the architecture and training objectives of the vision-language model remain unchanged.

We use the BLIP-2 ~\cite{li2023blip} as the vision and language backbone model. Given the recent success and efficiency of visual instruction methods, ~\cite{liu2023visual, zhu2023minigpt, li2023blip, dai2023instructblip}, we freeze the weights of visual and language model and only train the Qformer~\cite{liu2023visual} learns to map visual to text tokens. For discriminative tasks, we apply the stage 1 pre-training objective with Image-Text Contrastive, Image-Text Matching, and Image-Text Grounding Losses. We further explore generative performance with the FlanT5$_\mathrm{XXL}$ ~\cite{Wei2021FinetunedLM} language model and Mini-GPT4 that tunes the Vicuna-13b-v0 language model ~\cite{vicuna2023, Touvron2023LLaMAOA} to understand visual tokens. We refer to ~\cite{li2023blip} for more training details.




\section{Experiments \& Results}

\label{sec:experiments}

We use the OpenAI Chat API with gpt-3.5-tubro engine and a temperature of 0.8 to prompt the LLM to collect knowledge data. The BLIP-2 critic model is trained with total batch size of 256, learning rate of 1e-5, max 10 epochs. The visual encoder (ViT-G) model is additionally trained instead of kept it as frozen.

The discriminative BLIP2 is trained with 256 batch size and 128 max sequence length for 1e4 iterations. The BLIP-2 FlanT5$_\mathrm{XXL}$ and Mini-GPT4 models are trained with 64 batch size and 2e4 iterations. All models are trained with learning rate of 1e-5, Adam optimizer \cite{kingma2014adam}, linear warmup with cosine annealing, and image size of 480 using 80GB 4 A100 GPUS. We do not finetune the ViT or the language model, and only train the QFormer shown by the success from prior work ~\cite{li2023blip, dai2023instructblip, liu2023visual}.

\subsection{Downstream Tasks}

\paragraph{Localized Visual Commonsense Reasoning}
We evaluate on a set of visual commonsense reasoning tasks that involve identifying and referring specific regions in the image in a \textit{zero-shot} setting.  VCR~\cite{zellers2019recognition} is a task that requires choosing the right answers for question (Q $\to$ A), and rationales justifying the answer (QA$\to$ R) from four multiple choice options. The results are combined with (Q $\to$ AR) metric that requires selecting the right answer and rationale. VisualCOMET ~\cite{park2020visualcomet} is a commonsense knowledge graph of understanding specific people's intent, and what they would do before and after, and adopt their Acc@50 task of retrieving ground truth inferences from 50 candidates . Sherlock ~\cite{hessel2022abduction} is a visual abductive dataset that includes the comparison evaluation of ranking of 10 text inference candidates aligned with human preference. All the aligned tasks require reasoning about specific regions or people in the image, and getting the image-text similarity score from a model. 
\paragraph{Non-Localized Visual Reasoning}
We measure the effectiveness of the localized knowledge corpus on other vision-language tasks not limited to datasets with no bounding box annotations. We specifically focus on ones that require high-level reasoning that would benefit from visual commonsense corpus. AOKVQA \cite{schwenk2022aokvqa} requires outside world-knowledge to answer questions and we evaluate on their multiple choice setting.
SNLI-VE \cite{xie2019visual} is an inference based visual entailment that tests fine-grained image understanding. The task is to predict whether the image semantically entails the text, and specifically classify if the image-text is one of \{entailment, neutral, contradiction\}.  Visual7W ~\cite{zhu2016visual7w} is visual QA with focus on visual grounding, and we evaluate on the subset of telling questions that have textual answers (Telling QA).

\begin{table*}[t]
\normalsize
\centering
\scalebox{0.7}{
\begin{tabular}{l | c c c  c  c | c  c  c}

& \multicolumn{5}{c}{\textbf{Localized}} & \multicolumn{3}{c}{\textbf{Non-Localized}} \\
\toprule
\textbf{} & \multicolumn{3}{c}{\textbf{VCR}} & \multicolumn{1}{c}{\textbf{Sherlock}} & 
{\textbf{VisualCOMET}} &  
{\textbf{AOKVQA}} & 
{\textbf{SNLI-VE}} &
{\textbf{Visual 7w}}
\\
\textbf{Approach} & Q $\rightarrow$ A & QA $\rightarrow$ R & Q $\rightarrow$ AR & Comparison & Acc@50 & Mult. Choice & Classification & Telling QA \\ 
\midrule
CLIP-Event \cite{Li2022CLIPEventCT} & 52.4  & 49.2 & - & - & 22.4 & - & - & - \\ 
CLIP ViT-B-16$^*$ \cite{Radford2021LearningTV} & 54.8 & 48.6 & 26.6 & 9.9 & 33.0 & 58.3 & \textbf{36.0} & 65.9 \\
CLIP ViT-L-14x336 \cite{Radford2021LearningTV}  & \textbf{56.3} & \textbf{51.3} & \textbf{29.9} & 10.9 & 34.8 & 61.0 & 31.9 & 66.7 \\
BLIP ViT-L \cite{li2022blip} & 47.2 & 42.5 & 20.1 & 18.6 & 31.3 & 61.3 & 34.2 & 69.4 \\

BLIP-2 ViT-L \cite{li2023blip} & 52.3 & 48.1 & 25.3 & 18.7 & 36.7 & 65.0 &  31.7 & 73.6 \\
BLIP-2 ViT-G \cite{li2023blip} & 56.1 & 49.8 & 28.0 & \textbf{19.5} & \textbf{39.0} & \textbf{68.0} & 33.4 & 77.1 \\ 
\midrule 
BLIP-2 ViT-G + \method & \textbf{59.0} & \textbf{56.4} & \textbf{33.4} & \textbf{29.7} & \textbf{40.3} & \textbf{68.9} & \textbf{40.3} & \textbf{79.5}\\
\bottomrule
\end{tabular}
}
\caption {
   Zero-shot results on the localized and non-localized visual reasoning tasks. $^*$Zero shot VCR results directly obtained from \cite{Wang2022MultimodalAD}. For CLIP, we follow ~\cite{Wang2022MultimodalAD} by omitting the question and having the answer (with rationale) as text input to  calculate the image-text similarity. For BLIP-2, we maintain the question text input as it improves the performance. 
} 
\label{tab:downstream}
\vspace{0.5mm}
\end {table*}

\begin{figure*}[thb!]
\centering
  \includegraphics[width=0.8\textwidth]{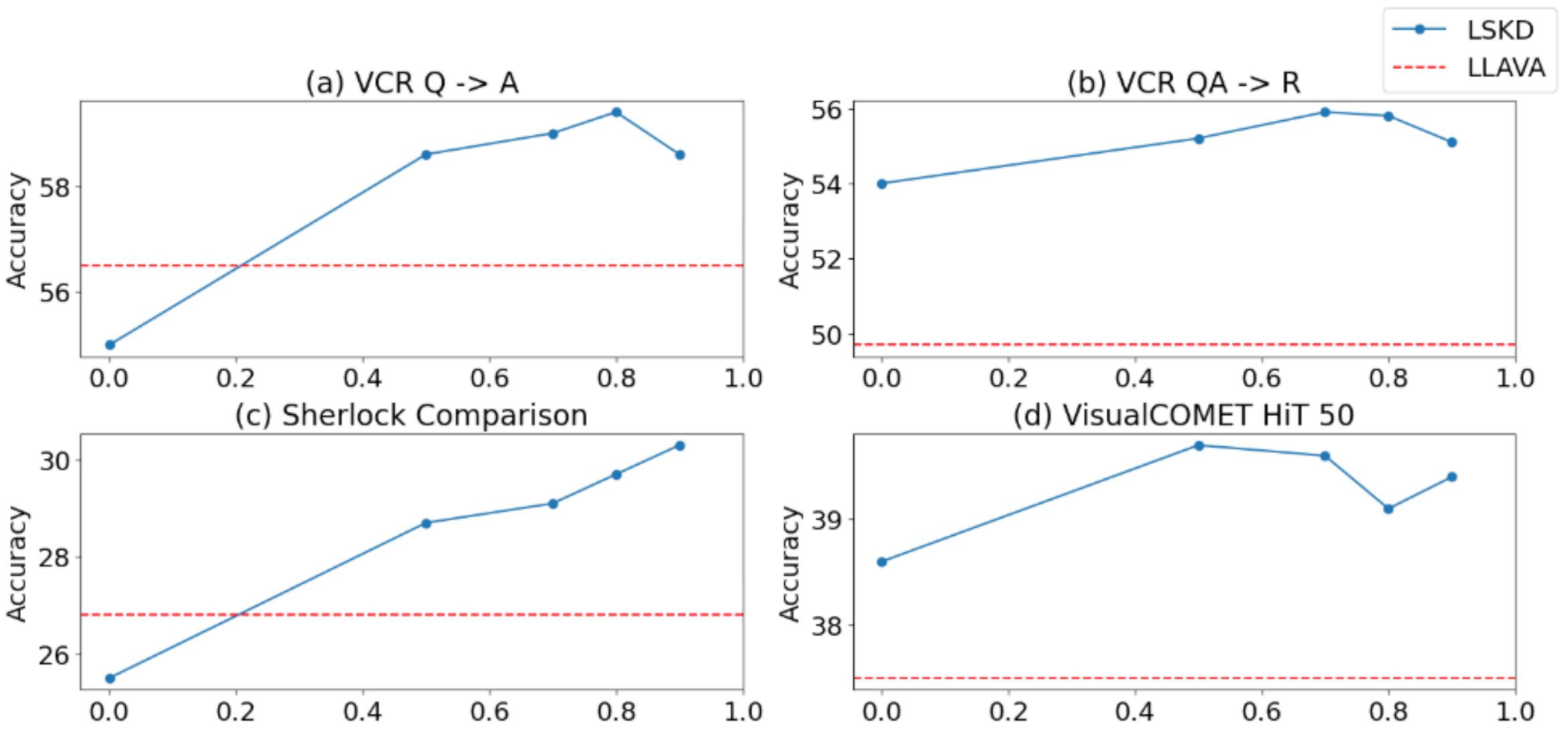}
{
  \caption{Effect of data quality controlled by filtering threshold on different datasets. The x-axis shows the threshold for filtering and the y-axis is the accuracy metric in percentage. We compare training our model on the LLaVA-instruct dataset (red) and ours (blue).}%
  \label{fig:thresholding}
}

\end{figure*}

\paragraph{Baseline models} We include CLIP as our baseline as it has shown strong zero-shot generalization results for various image-text alignment tasks ~\cite{Radford2021LearningTV}. Following \cite{Wang2022MultimodalAD}, we exclude the question in the text input and acquire the image-text cosine similarity score to do the task. CLIP-Event is a CLIP model pre-trained on their VOA dataset crawled from news websites ~\cite{Li2022CLIPEventCT}. BLIP is image-text alignment model trained with additional generation objective and boostrapped image captions ~\cite{li2022blip}. We lastly evaluate the zero shot performance of BLIP-2 ~\cite{li2023blip} varying the visual encoders before applying knowledge distillation. We do not draw bounding boxes in the image nor include id tags in the text description, as these models have not been pre-trained in this domain. 

\subsection{Zero-Shot Visual reasoning results}
Table~\ref{tab:downstream} shows the zero-shot results on the downstream tasks. For localized reasoning tasks, we first observe that scaling the visual encoder size (CLIP ViTB-16 vs ViT-L-14x336; BLIP-2 ViT-L vs ViT-G) in general improves the performance. CLIP outperforms BLIP-2 on VCR tasks but fall short on Shlerock and VisualCOMET.  After applying localized symbolic knowledge distillation (LSKD) to BLIP-2, there is a consistent improvement over the BLIP-2 model on all downstream tasks (5.4\% on VCR Q $\to$ AR, 10.2 on Sherlock Comparison, 1.3\% on VisualCOMET Acc@50).

For non-localized reasoning tasks, we observe a similar pattern. Interestingly, applying LSKD improves the performance of BLIP2 model further across all the tasks (AOKVQA, SNLI-VE, Visual7W) over the vanilla model, despite these tasks not being the primary target domain. This demonstrates that the advantages of distilling models with localized reasoning can be transferred to high-level visual commonsense tasks, thanks to the visual precision and enhanced reasoning abilities learned from the generated knowledge corpus. 






\begin{table*}[t]
\normalsize
\centering
\scalebox{0.7}{
\begin{tabular}{l l c | c  c  c  | c  c c}
& & & \multicolumn{3}{c}{\textbf{Localized}} & \multicolumn{3}{c}{\textbf{Non-Localized}} \\
\toprule
\textbf{} & & &  \textbf{VCR} & \multicolumn{1}{c}{\textbf{Sherlock}} & 
{\textbf{VisualCOMET}} &  
{\textbf{AOKVQA}} & 
{\textbf{SNLI-VE}} &
{\textbf{Visual 7w}}
\\
\textbf{Dataset} & \textbf{Size} & \textbf{Annotator} & Q $\rightarrow$ AR & Comparison & Acc@50 & Mult. Choice & Classification & Telling QA \\ 
\midrule
Zero-Shot & NA & NA & 28.0 & 19.5 & 39.0 & 68.0 & 33.4 & 77.1 \\
\midrule
Sherlock \cite{hessel2022abduction} & 300K & Human &  34.6 & 30.5 & 39.7 & 67.2 & 38.6 & 70.1 \\
VisualCOMET \cite{park2020visualcomet} & 1.2M & Human & 31.8 & 25.3 & 50.2 & 68.5 & 35.6 & 70.8 \\
\midrule
LLAVA-Instruct \cite{liu2023visual} & 150K & GPT-4 & 28.1 & 26.9 & 37.5 & 71.0 & 42.6 & 79.5 \\
\method (Ours) & 150K & ChatGPT &  33.3 & 28.6 & 39.7 & 69.6 & 38.0 & 75.9 \\
\method (Ours) & 1M & ChatGPT & 33.4 & 29.7 & 40.3 & 68.9 & 40.3 & 79.5 \\

\bottomrule
\end{tabular}
}
\caption {
    Ablations of BLIP-2 ViT-G  trained with varying sources of visual-knowledge corpus annotated by humans and machines. We break down to visual reasoning tasks that require localized reasoning and those do not. Critic filtering is applied to the LSKD corpus (Ours).
} 
\label{tab:filtering_ablations}
\vspace{0.5mm}
\end {table*}

\paragraph{Influence of Critic Filtering on Downstream Tasks}

How does the process of critic filtering influence the performance of downstream tasks? Keeping the size of the selected statements the same at $\sim$ 300K, we select qualified knowledge statements with varying prediction thresholds. We also compare with training on the LLaVA-instruct dataset which similarly prompts an LLM (GPT-4) to generate complex questions using ground truth verbalizers ~\cite{liu2023visual}. Figure ~\ref{fig:thresholding} presents the resulting performances at these diverse thresholds across different datasets. Compared to LLaVA, we observe that localized knowledge statements without filtering does not show any improvement for the downstream model, while any thresholding over 0.2 is consistently better than LLaVA across all datasets. For tasks that demand relatively moderate commonsense, such as VCR Q$\rightarrow$A and Sherlock Comparison, increasing the threshold consistently improves the model performance. For tasks requiring a higher degree of commonsense such as VCR QA$\rightarrow$R and VisualCOMET Hit@50, the performance increases until a certain threshold and then fluctuates. 
We speculate that a more grounded critic model could potentially mitigate this fluctuation, and we intend to investigate this in our future work. Overall, our findings suggest that higher thresholds (i.e., more critical filtering) tend to yield superior quality generations, thereby enhancing the performance on downstream tasks.

\subsection{Human vs Machine Annotated Corpus}
Can training on machine annotated corpus result in competitive performance with human annotations? 
In Table~\ref{tab:filtering_ablations}, we compare the performance of BLIP-2 ViT-G trained on existing human-annotated corpora with our machine-annotated corpus across various scales. First, we found that increasing the size of our training corpus (150K vs 1M) leads to consistent improvement across all tasks, indicating a promising trend of scaling law for synthetic training data. Regardless of the size, training on our dataset yields considerable benefits over the zero-shot model on localized reasoning tasks.  

Next, we observe that training on human annotated corpus vastly improves the performance of their relative tasks (e.g. training on VisualCOMET boosts performance from 39.0 to 50.2). However, this can lead to inferior results on other visual reasoning tasks than the zero-shot counterpart. For instance, the performance on Visual7W drops from 77.1 (Zero-shot) to 70.1 (Sherlock) and 70.8 (VisualCOMET). This suggests that human-designed datasets may limit task generalization due to their lack of diversity. Interestingly, we see that training the model our full LSKD corpus (1M) leads to uniform gains over the zero-shot model across the tasks, and even outperforms the human annotation corpus for the non-localized tasks as well. This shows that machine-annotated datasets, when curated and scaled adequately, can indeed rival or even surpass the performance of models trained on human-annotated corpora.

We directly compare training on ours and the LLaVA dataset. Regardless of our dataset scale, we observe that LSKD + filtering wins over training on the LLaVA corpus on localized reasoning benchmarks, even when using a less powerful teacher model (ChatGPT vs GPT-4). This suggests that our creation of a new localization corpus is crucial to support the model with grounded reasoning. On the other hand, LLAVA wins on non-localized reasoning tasks as they are aligned with the nature of training corpus. We thus observe that the appropriate application of the corpus can be task-dependent, and adopting a selective approach towards generating the corpus may result in significantly enhanced performance across various benchmarks.

\subsection{Localized Reasoning with Generative Models}
\label{sec:generative}
We extend \method to train generative models that can refer and talk about highlighted regions in image. We finetune BLIP-2 FlanT5 and Mini-GPT4 and prompt them to answer questions from the VCR data. As there is no baseline zero-shot model that can reason about regions to answer questions,  we make a direct comparison of the student \method model to the teacher LLM with access to verbalizations. 
We ask annotators on Amazon Mechanical Turk (AMT) platform to run head-to-head comparisons (with ties) on three criteria, if the answer delivers: 1) visually correct details, 2) informative and interesting information, and 3) content that sounds plausible. Finally, they select their overall preference. We take the majority vote of 3 annotators, and disregard the instance if there is no clear majority. 

\begin{table}[t]
\normalsize
\centering
\scalebox{0.8}{
\begin{tabular}{l| c c c c}
\toprule

\textbf{Model} & \multicolumn{1}{c}{\textbf{Correctness}} & \multicolumn{1}{c}{\textbf{Informativeness}} & \multicolumn{1}{c}{\textbf{Plausibility}} & \multicolumn{1}{c}{\textbf{Overall}}  
\\ 
\midrule 
ChatGPT w/ Vebalizers & 34.7 & 33.9 & 39.6 & 45.0 \\
BLIP-2 (FlanT5$_\mathrm{XXL}$-11B) + \method & 31.7 & 41.0 & 30.2 & 41.2 \\
\multicolumn{1}{c}{Tie} & 33.7 &  25.1 & 30.2 & 13.1 \\

\midrule
ChatGPT w/ Vebalizers  & 29.8 & 31.7 & 36.8 & 40.6 \\
Mini-GPT4 (Vicuna-13B) + \method & 34.3 & 53.0 & 34.2 & 49.1 \\
\multicolumn{1}{c}{Tie} & 35.9 & 15.3 & 30.0  & 10.3 \\
 \bottomrule
\end{tabular}
}
\caption {
Human evaluation of generative models with \method vs Chat-GPT with verbalizers. Humans are asked to choose the better generation or tie if they share the same quality.
} 
\label{tab:human_eval}
\end {table}

Table ~\ref{tab:human_eval} shows the human evaluation results. We observe that the \method generally wins in informativeness over ChatGPT, but not in plausibility. We see a conflicting pattern in correctness and overall preference, where Mini-GPT4 is equipped with a more powerful language model that outperforms the teacher model while BLIP-2 falls short. Unlike previous language-based distillation where a relatively weak student model can outperform the teacher \cite{west2021symbolic, Bhagavatula2022I2D2IK}, we see that a strong student model may be required to outperform the teacher LLM in the multimodal domain.   

\begin{figure*}[t]
\centering
\includegraphics[trim=0cm 0cm 0cm 0cm, width=0.99\linewidth]{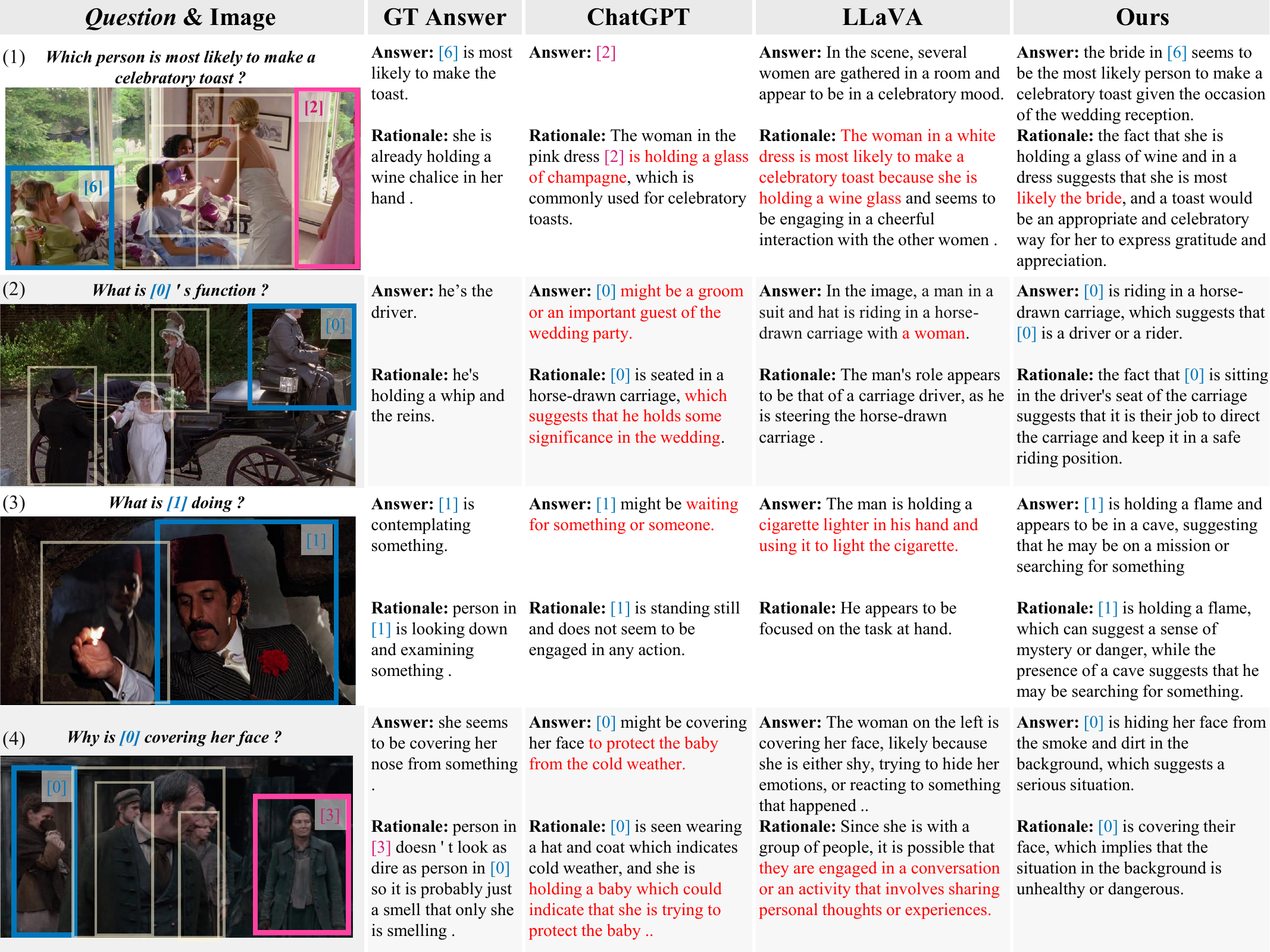}
\caption{Qualitative examples comparing ChatGPT (the teacher model), LLAVA trained for complex visual reasoning \cite{liu2023visual}, and ours. Each person referenced in the question has been assigned a unique number with a bounding box and their mention has been marked with a corresponding color. Any errors within the generated results are emphasized with a red highlight.}
\label{fig:qual-examples}
\end{figure*}

\paragraph{Qualitative Results}
Figure~\ref{fig:qual-examples} presents a comparative analysis of question-answering with rationale results on VCR samples generated by ChatGPT, LLaVA~\cite{liu2023visual} and \texttt{Ours}.
Both Ground Truth (GT) and \texttt{Ours} consistently identify the correct entities, with \texttt{Ours} model often providing broader context, which is uncertain on rare occasions (\eg ``likely the bride''). On the other hand, ChatGPT predominantly focuses on observable actions or states as described in the text context, occasionally resulting in the misidentification of the visual entities and their relations. In the third example in Figure~\ref{fig:qual-examples}, ``waiting for someone'' focuses on the observable state ``standing still'', missing visual detail such as a cave, holding a flame, and surrounding context.
LLaVA, in contrast, generally provided a broad overview while identifying a specific visual entity in most cases. However, it often struggled to accurately identify specific entities within the complex scene (\eg ``holding a wine glass'' in Figure~\ref{fig:qual-examples}.(1), ``cigarette'' in Figure~\ref{fig:qual-examples}.(3) ). Compare to LLaVA, \texttt{Ours} often aligned closely with GroundTruth and incorporated both actions and inferred knowledge in its answer. Overall, \texttt{Ours} delivers a more nuanced and contextually-rich response.

\vspace{-3mm}
\section{Related Work}
\vspace{-1mm}
\paragraph{Knowledge Distillation}

Recent research \cite{DBLP:journals/corr/abs-2204-06031} has extensively explored the use of language models as knowledge bases, highlighting their potential in reasoning, explainability, and consistency, which can enhance downstream tasks by distilling knowledge from LMs.
\cite{DBLP:conf/naacl/GuiWH0BG22} demonstrated how knowledge augmentation explicitly from knowledge bases and implicitly from GPT-3 improved open-domain multimodal tasks. 
\cite{DBLP:conf/emnlp/LiuSSC22} showed that overgeneration with GPT-3 from exemplars to filter, as well as reviewed by humans, is a new and reliable way to create an NLI dataset with human and AI collaboration. This setup also has the advantage of bringing forth cultural internalization via human collaboration \cite{DBLP:journals/corr/abs-2206-01134}. Previous works have explored knowledge distillation in the multimodal domain by prompting the teacher LLM with human-annotated verbalizations \cite{liu2023visual, zhu2023minigpt, dai2023instructblip}. Our work is different in that it generated \textit{localized} commonsense descriptions and the knowledge generation framework can operate a scale without the need for aligned descriptions. 


\paragraph{Filtering}

\cite{DBLP:conf/aaai/Anaby-TavorCGKK20} filters the generated sentences using a classifier trained on original sentences and a set of generated sentences. \cite{DBLP:conf/acl/0003XSHTGJ22} used the same technique to filter out synthetic data created, which is of low quality.
Large language models can be used to refine the commonsense knowledge retrieved from web contents by filtering the data generated from these models \cite{DBLP:journals/corr/abs-2112-04596}. They perform a consolidation step that filters topical and relevant assertions based on OpenIE.


\paragraph{Multimodal commonsense reasoning} requires more profound real-world knowledge, potentially spanning logical, causal, and temporal relationships between concepts. For example, elements of causal reasoning are required to answer the questions regarding images in VCR~\cite{zellers2019recognition} and VisualCOMET~\cite{park2020visualcomet}, while other works have also introduced datasets with video and text inputs to test for temporal reasoning (\eg, Social-IQ~\cite{zadeh2019social}, MovieQA~\cite{tapaswi2016movieqa}, MovieFIB~\cite{maharaj2017dataset}, TVQA~\cite{lei2018tvqa}). Benchmarks for multimodal commonsense typically require leveraging external knowledge from knowledge bases~\cite{song2021kvl} or pretraining paradigms on large-scale datasets~\cite{lu2019vilbert,zellers2021merlot}.

\paragraph{Region Understanding with Multimodal alignment}

Capturing elements across modalities that have a common meaning and is exemplified by tasks such as visual coreference resolution~\cite{kottur2018visual,park2020identity}, visual referring expression recognition~\cite{cirik2018using}, multimodal question answering~\cite{hessel2022abduction,zellers2019recognition}, and cross-modal retrieval~\cite{frome2013devise,plummer2015flickr30k}. Alignment between modalities is challenging since it may depend on long-range dependencies, involves ambiguous segmentation (\eg, words or utterances), and could be either one-to-one, many-to-many, or not exist at all. Resources for fine-grained alignment include Visual Genome~\cite{krishna2017visual} and dense captioning~\cite{johnson2016densecap}, diverse reasoning  ~\cite{yin2021broaden}.  Recent methods have adopted either generative or retrieval-based methods  for alignment: generative methods create localized verbalizations of the region of interest ~\cite{zhou2020more,fu2016aligning,johnson2016densecap, zhong2022regionclip}, while retrieval aims to select the most accurate caption for the region of interest despite possibly given only coarse-grained paired data of captions for entire images ~\cite{chen2020fine,he2019new}.









\section{Conclusion}

We present \method, a method for sampling localized commonsense knowledge from a large language model. With the help of a supervised critic model aligned with human judgments, we create a diverse, reliable 1M localized commonsense corpus. Training on the resulting corpus supports models that can accept region references as input, which allows users to interact with specific parts of images by ``pointing;" all without having to write out a referring expression explicitly. 
We show that training on our corpus improves the zero-shot performance of vision-language models for tasks requiring regions-as-input. Making the critic model more critical by strict thresholding improved performance further. We present a state-of-the-art zero-short performance with our approach opening avenues for visual commonsense models with our localized commonsense knowledge corpus.

\section*{Acknowledgements}
We thank members of the Mosaic team at AI2 and Microsoft Research Deep Learning team for valuable discussions. This research was supported by the NSF (DMS-2134012, IIS-1652052, and IIS-1703166],  DARPA MCS program through NIWC Pacific (N66001-19-2-4031), and the Allen Institute for AI.

\bibliographystyle{plain}

\clearpage
\appendix
\section*{Supplementary Material}

\section{The Localized Commonsense Knowledge Corpus}
\label{sec:corpus_detail}
\begin{figure}[htb!]
\begin{floatrow}

\raisebox{5ex}{
\ffigbox[\FBwidth]
  {\includegraphics[width=0.25\textwidth]{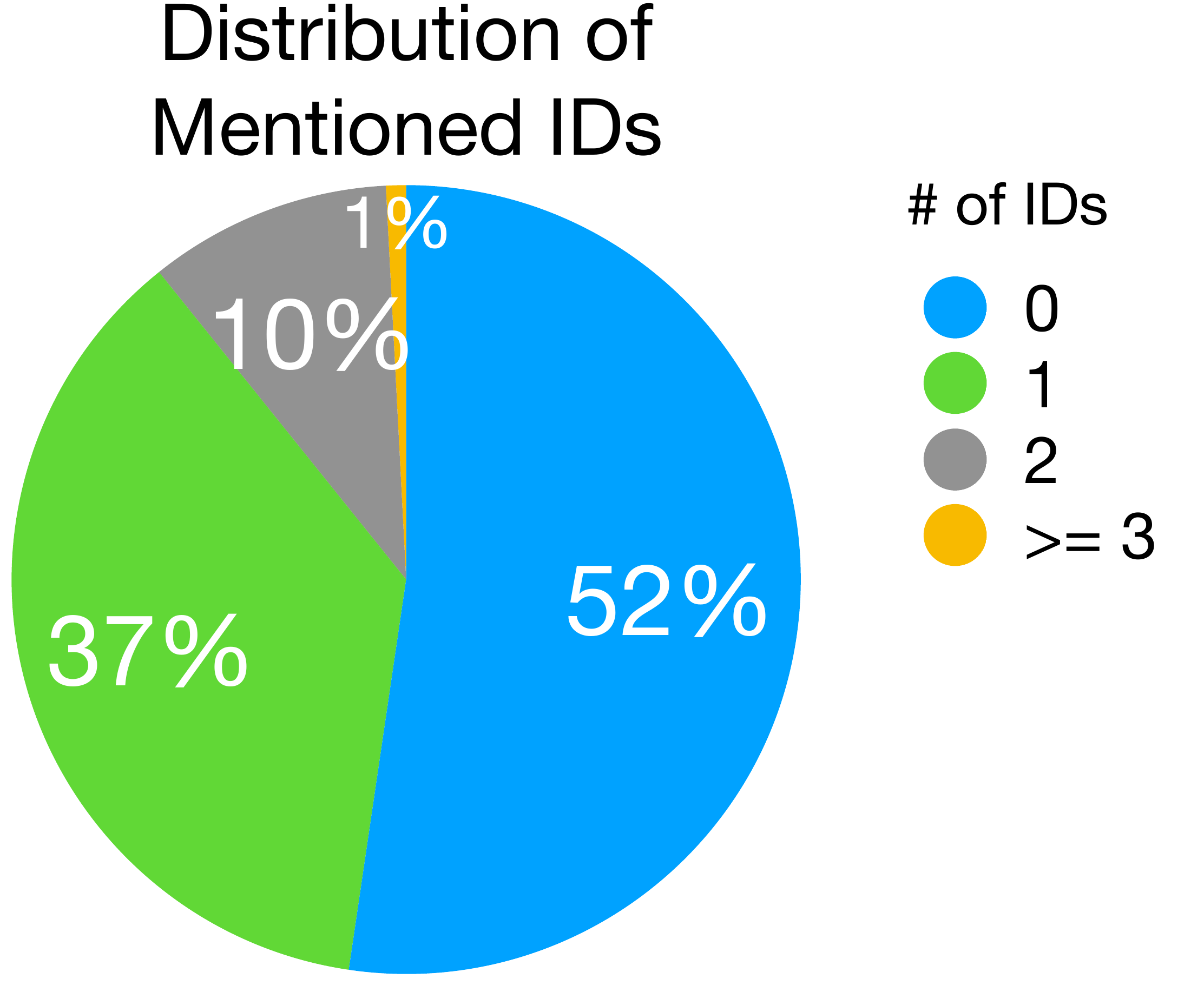}}
  {\caption{Distribution of mentioned Region ID's.}
  \label{fig:id_distribution}
  }
}
  
\capbtabbox
{
\normalsize
\centering
\scalebox{0.75}{

\begin{tabular} { l c l }
\toprule

\multicolumn{1}{c}{\textbf{Question Type}} & \multicolumn{1}{c}{\textbf{Freq (\%)}}  & \multicolumn{1}{c}{\textbf{Example}} \\
\midrule

Purpose & 20.0 & \textit{What is the purpose, What is the significance...}\\
Relationship & 10.5 & \textit{What is the relationship, How are they related...} \\
Type & 10.1 & \textit{What kind of, What is the type of... } \\
Emotion & 8.4 & \textit{What emotion, What might be the feeling of...}\\
Scene & 7.7 & \textit{Where, What time, What situation...}\\
Attribute & 7.4 & \textit{What state, What condition, What color...} \\
Action & 5.9 & \textit{What activity, What event, What are they doing...} \\
Inference & 5.3 & \textit{What can you infer, What would likely, How might...} \\
Reason & 5.1 & \textit{Why, What is the intention...} \\
Role & 4.7 & \textit{What is the role, What is the occupation...}\\
Focus & 4.5 & \textit{What is the main focus, What stands out...}\\
Ambiance & 4.4 & \textit{What atmosphere, What is the mood, What vibe...}\\
Factual & 3.5 & \textit{Is/Are there..., Do you think...} \\
Others & 2.6 & - \\
\bottomrule
\end{tabular}
}
\caption {
Types of questions and their examples in the corpus.  To identify these question types, we manually construct a set of n-gram patterns and categorize questions based on their inclusion of these specific n-grams.  
} 
\label{tab:question_types}
\vspace{0.5mm}
}


\end{floatrow}
\end{figure}

Table \ref{tab:dataset} shows the detailed statistics of the corpus. We break down the corpus where the regions are referenced by their IDs and by their region descriptions. The maximum number of mentioned region IDs in the QAR is limited to 5. Figure \ref{fig:id_distribution} illustrates the distribution of the number of IDs.

\begin{figure}[htb!]
  \centering
  \begin{subfigure}{0.32\linewidth}
    \centering
    \includegraphics[width=\linewidth]{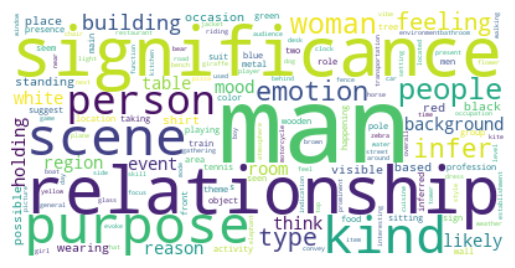}
    \caption{Question}
    \label{fig:question_wc}
  \end{subfigure}%
  \begin{subfigure}{0.32\linewidth}
    \centering
    \includegraphics[width=\linewidth]{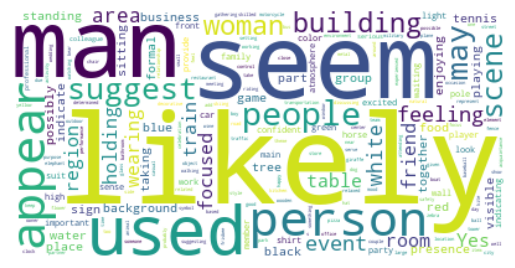}
    \caption{Answer}
    \label{fig:answer_wc}
  \end{subfigure}%
  \begin{subfigure}{0.32\linewidth}
    \centering
    \includegraphics[width=\linewidth]{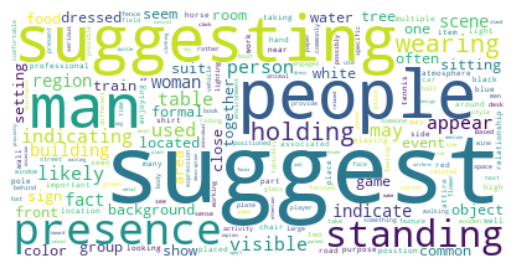}
    \caption{Rationale}
    \label{fig:rationale_wc}
  \end{subfigure}
  \caption{Word Clouds of Question, Answer, and Rationale}
  \label{fig:word_cloud}
\end{figure}

We show the category of question types and examples in Table ~\ref{tab:question_types}. Upon manual inspection of the corpus, we have identified specific question types that exhibit prominent characteristics. These types are associated with a collection of n-gram patterns, and questions sharing these n-grams are categorized accordingly (e.g., questions containing the terms "purpose" and "significance" are assigned to the Purpose category). Lastly, the word clouds for question, answer, and rationale are shown in Figure~\ref{fig:word_cloud}.

\begin{table}[htb!]
\normalsize
\centering
\scalebox{0.99}{
\begin{tabular}{l| c c c }
\toprule

\textbf{} & \multicolumn{1}{c}{\textbf{With Region ID's}} & \multicolumn{1}{c}{\textbf{With Region Descriptions}}  & \multicolumn{1}{c}{\textbf{Total Corpus}}
\\ 
\midrule 
\# of Images & 128,564 & 125,524 & 168,996 \\
\# of QARs & 513,223 & 467,658 & 1,023,807 \\ 
Average \# of Qs per Image & 3.99 & 3.73 & 3.86 \\ 
\midrule
Average Q Length & 13.0 & 10.9 & 11.8 \\
Average A Length & 14.4 & 10.5 & 12.3 \\ 
Average R Length & 25.8 & 22.8 & 24.1 \\
Average \# of mentioned ID's & 0 & 1.25 & 0.57 \\
 \bottomrule
\end{tabular}
}
\caption {
Detailed statistics of the Localized Commonsense Knowledge Corpus.
} 
\label{tab:dataset}
\vspace{0.5mm}
\end {table}

\begin{figure*}[h!]
\centering
  \includegraphics[width=0.9\textwidth]{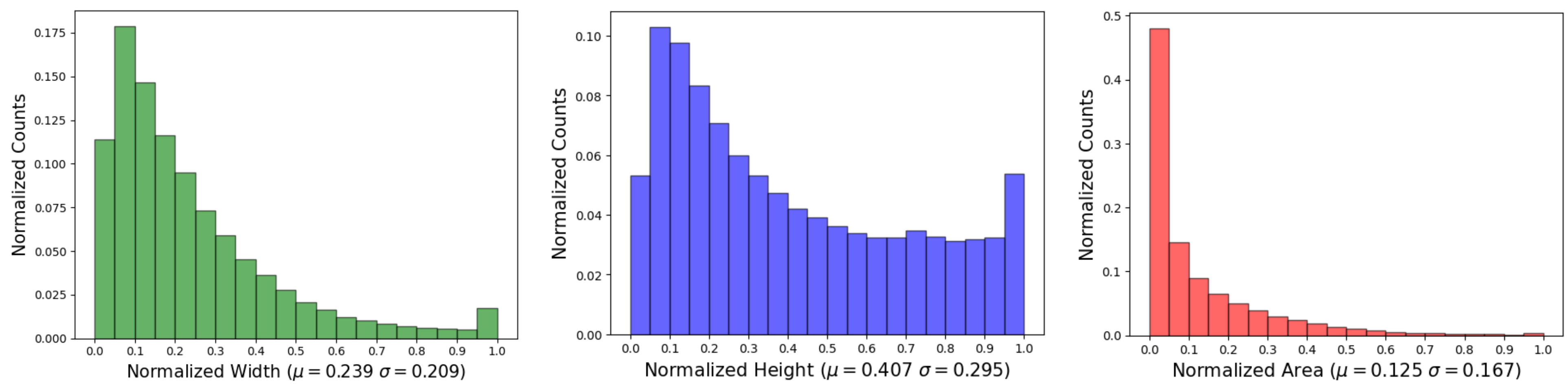}
{
  \caption{Distribution of bounding box sizes in the generated corpus. x-axis is the normalized box width, height, and area from 0 to 1. y-axis is the normalized counts over total number of samples.}%
\label{fig:bbox_distribution}
}
\end{figure*}

\section{Bounding Box Distributions and Model Performance}
Figure~\ref{fig:bbox_distribution} shows the distribution of normalized bounding box sizes in the filtered corpus, highlighting the width, height, and the area. We notice that almost 50\% of the bounding boxes have the normalized area 
 0.05, suggesting that small objects are well-covered in our corpus. The height shows more uniform distribution than the width, indicating that there are more bounding boxes with smaller widths and the width mainly clusters in the range of 0.1-0.2. This reveals that the corpus contains not just large and prominent objects, but also small or narrow objects that often require attentive vision models to recognize.

We use the Sherlock comparison task \cite{hessel2022abduction} to study the model performance change w.r.t different bounding boxes as their dataset consists of single bounding boxes with diverse sizes. The Pearson’s correlation between the input bounding box size and the comparison accuracy is $\rho= -0.12$ with p-value of 0.05.

Based on the correlation, we see that the performance is actually higher for smaller objects. One might indeed initially think that larger bounding boxes would result in better performance, as they could potentially encompass more features or more of the object of interest. We hypothesize that the negative correlation is due to the following.
\begin{itemize}
    \item{Specificity}: Smaller bounding boxes quite often are more specific in identifying the target objects, thereby reducing the complexity of the region and making it easier for the model to focus and reason.
    \item{Clutterness}: Larger bounding boxes might include more "noise" or irrelevant objects/background, which could mislead the model during the reasoning process as it gets distracted by extraneous details.
\end{itemize}

\section{More Details of Corpus Generation}
\label{sec:corpus_generation}
\begin{figure*}[htb!]
\centering
\includegraphics[trim=2cm 1cm 0cm 0cm, width=0.99\linewidth]{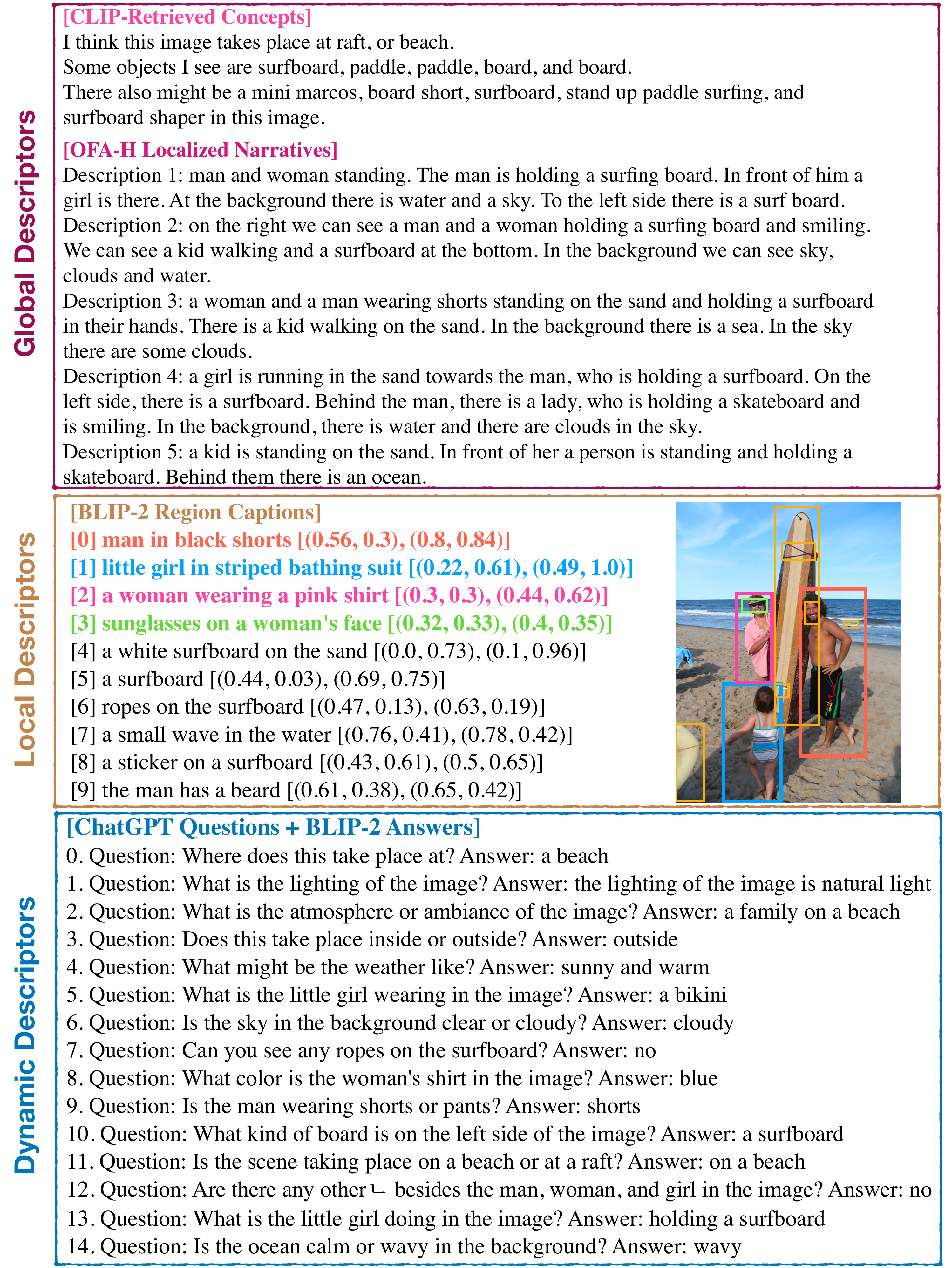}
\caption{Example of image-to-text verbalization with diverse descriptors.
}
\label{fig:verbalization}
\end{figure*}

We show the full output of our image to text verbalization pipeline using the global, region, and question-answer descriptors in Figure~\ref{fig:verbalization}. For concepts, we acquire the visual feature $v$ and text features for each object classes $[t_1, t_2, ...t_C ]$ extracted by the CLIP-ViT-L-336 model \cite{Radford2021LearningTV}, and use the nearest neighbor search by their cosine distance to select the top $k$ labels for the image. We train OFA-Huge model ~\cite{wang2022ofa} on the Localized Narratives ~\cite{PontTuset_eccv2020} and generate 5 descriptions with nucleus sampling ~\cite{holtzman2020curious} of $p=0.95$. BLIP-2 trained on region captions described in Section 2.1 is used to describe the regions individually. We get the questions using ChatGPT, in which we provide the global and local descriptors as context, and call the OpenAI API with the following instruction: \texttt{Here is the context for the image: \{global descriptors\} $\backslash \textrm{n} \backslash \textrm{n}$ \{local descriptors\} $\backslash \textrm{n} \backslash \textrm{n}$ Now, ask fifteen interesting but simple questions that you want to ask so you can get more understanding about the image}. The zero-shot BLIP-2 answers the generated question, and the QA pairs are used as the dynamic descriptors. 

\begin{figure*}[htb!]
\centering
\includegraphics[trim=2cm 1cm 0cm 0cm, width=0.99\linewidth]{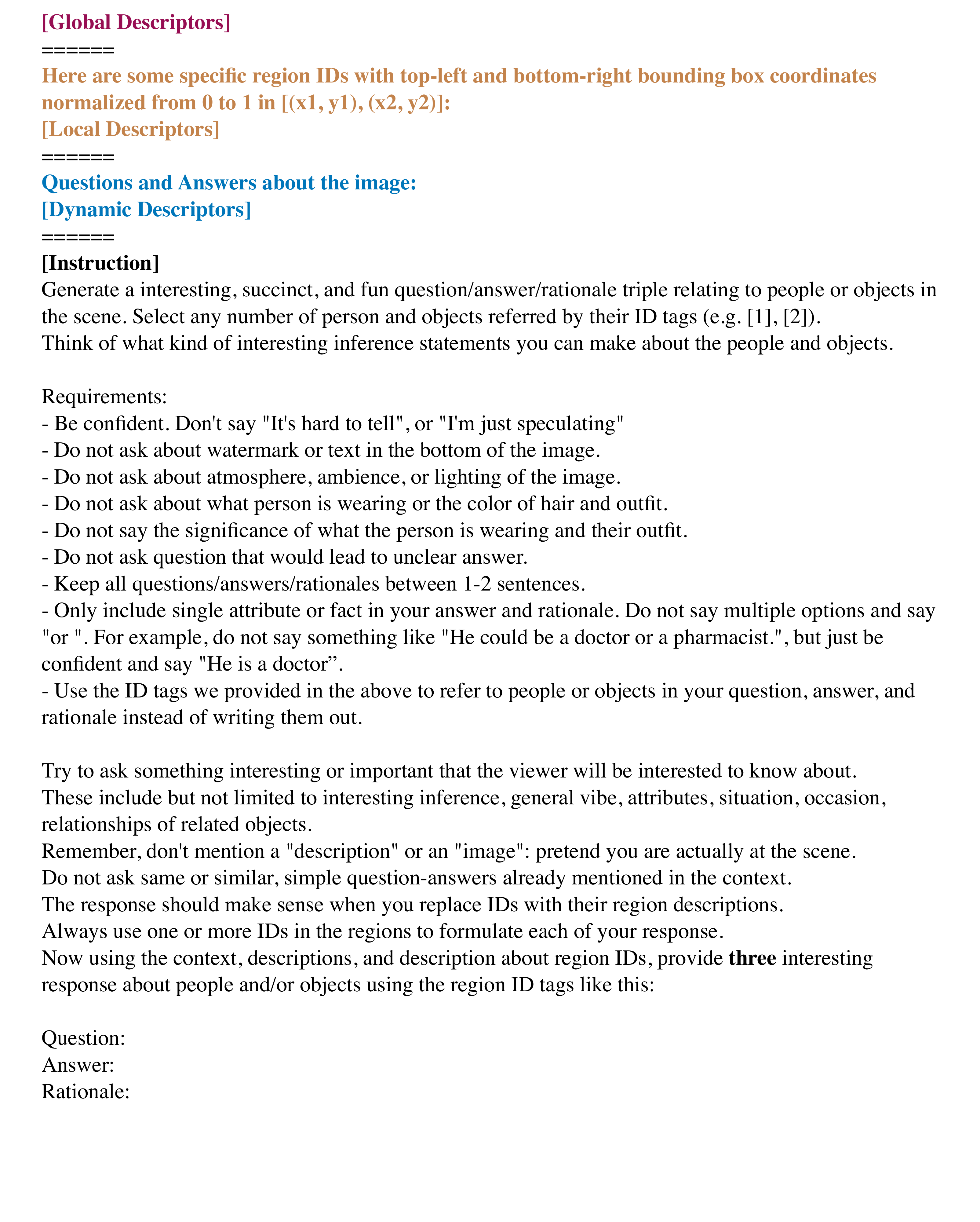}
\caption{Prompt used to generate data while referring regions by numerical IDs.
}
\label{fig:prompt1}
\end{figure*}

\begin{figure*}[htb!]
\centering
\includegraphics[trim=2cm 1cm 0cm 0cm, width=0.99\linewidth]{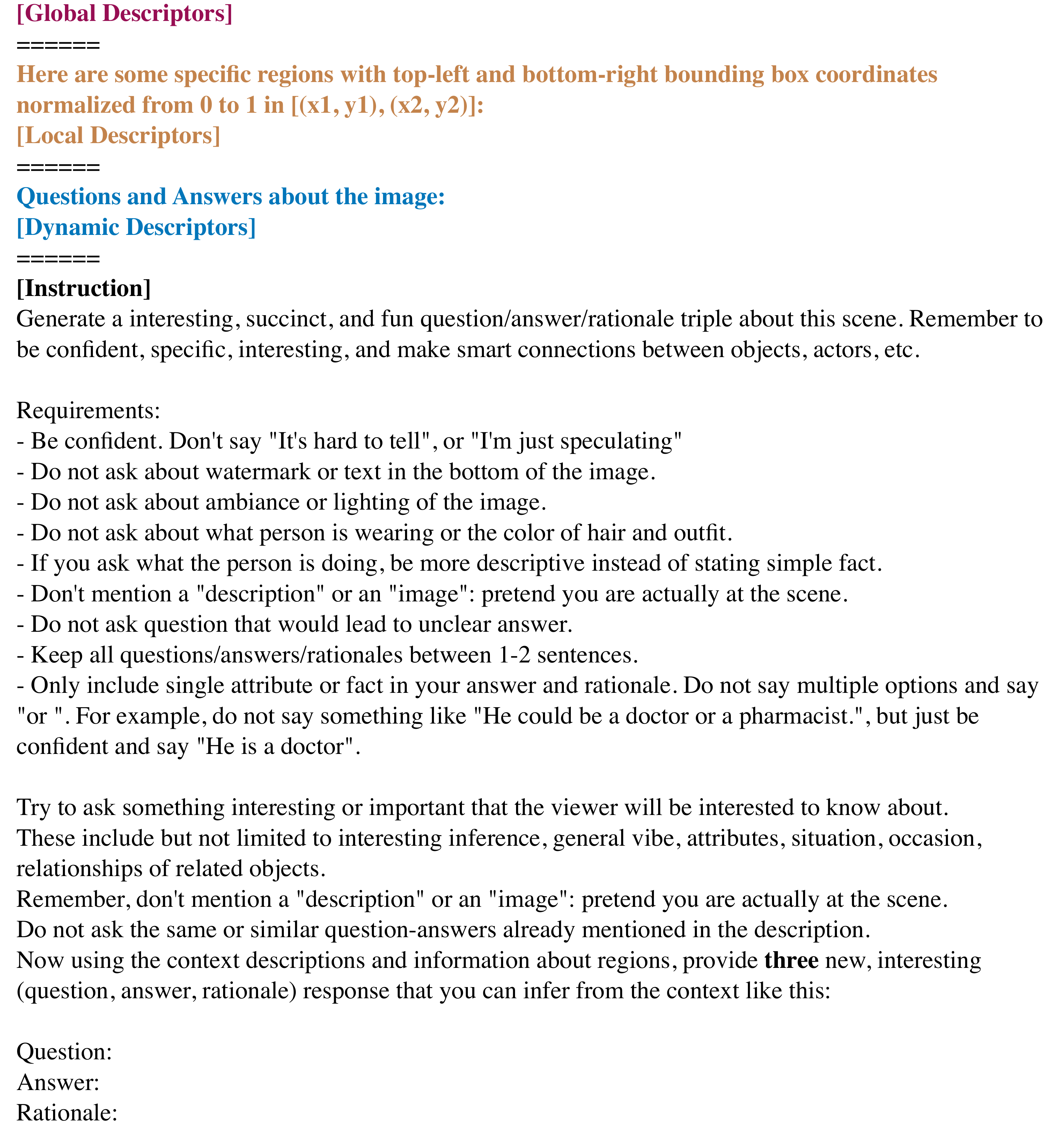}
\caption{Prompt used to generate data while referring regions by their descriptions.}
\label{fig:prompt2}
\end{figure*}

\begin{figure*}[htb!]
\centering
\includegraphics[trim=2cm 1cm 0cm 0cm, width=0.99\linewidth]{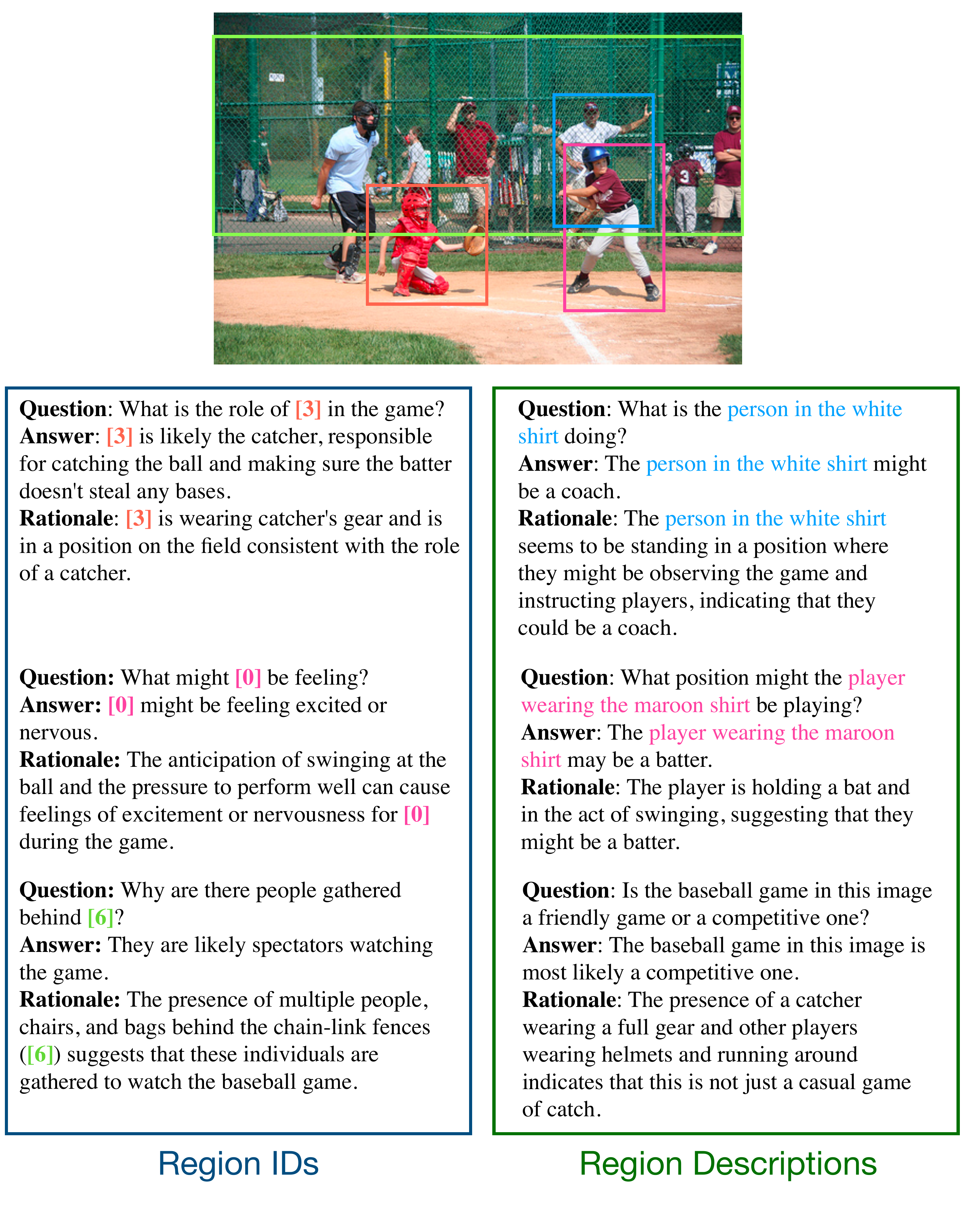}
\caption{Example of our generated corpus referring regions by IDs or descriptions.
}
\vspace{0.5cm}
\label{fig:corpus_example}
\end{figure*}

To generate the Localized Commonsense Knowledge Corpus, we utilize verbalization as context and present two distinct prompts to ChatGPT. In one prompt, regions are referenced by numerical IDs, while in the other prompt, regions are described using text descriptions. The specific prompts used to invoke ChatGPT are depicted in Figure \ref{fig:prompt1} and Figure \ref{fig:prompt2}. In the former case, instances where no IDs are mentioned in the output are filtered out, while in the latter case, instances containing any IDs in the output are excluded. An example generated using both versions of the prompts is showcased in Figure \ref{fig:corpus_example}.

\begin{figure*}[h]
\centering
  \includegraphics[width=0.75\textwidth]{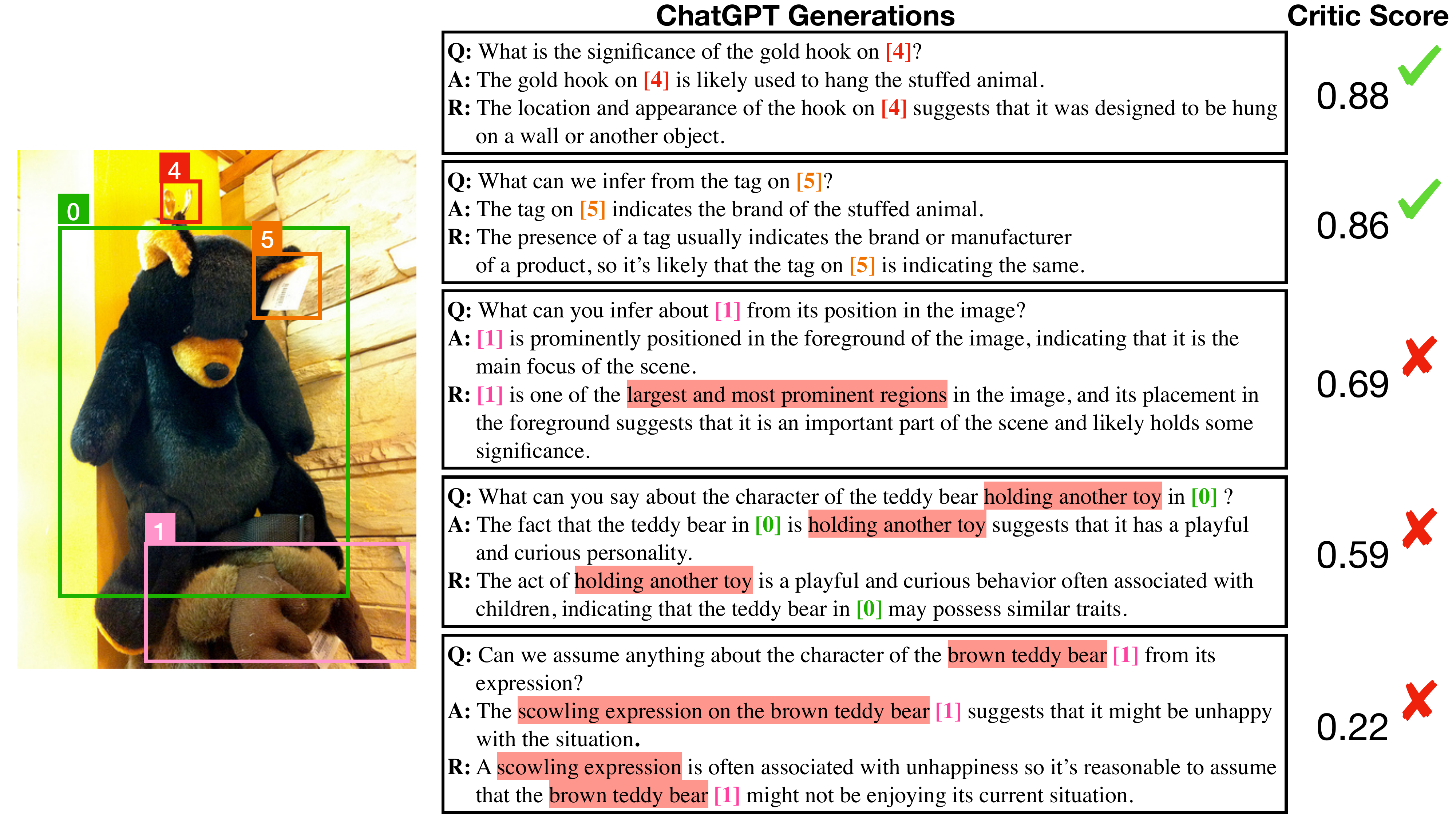}
{
    \caption{Qualitative examples of supervised critic filtering of ChatGPT generated data. We discard generations whose critic scores are lower than the threshold value of 0.8. Incorrect visual details are highlighted as red.}
    \label{fig:critic_qualitative}
}

\end{figure*}

\section{Qualitative Analysis of Critic Model}
Figure \ref{fig:critic_qualitative} shows qualitative examples to understand the patterns of critic model in distinguishing good and bad examples. We see that the model mostly relies on incorrect visual details (highlighted as red) lower than the correct instances. The third instance does not have glaring visual errors but are scored lower due to statement of "largest and most prominent regions", which is ambiguous but close to false. The critic model displays good calibrations in ordering the instances, such as giving the lowest score to the instance with the most visual error.

\section{More details of Local Descriptors}
\label{sec:local}
We train a region captioning model that maps from (image, region) $\rightarrow$ description of the region. We fine-tuned the generative version of BLIP-2 \cite{li2023blip} with the FLAN-t5-xxl \cite{chung2022scaling} backbone. We trained on a combination of
RefCOCO/RefCOCO+/RefCOCOg \cite{yu2016modeling,mao2016generation} (120K/80K/120K training region captions), Sherlock Clues-only \cite{hessel2022abduction} (277K), and VisualGenome \cite{krishna2017visual} (1.96M): all of these datasets provide descriptions of given regions within images. Following \cite{zellers2021merlot,yao2021cpt}, we render the bounding box in the image itself to allow the model access to the bounding box's location.

We compared our model's captions to those generated by GRiT \cite{wu2022grit}, which achieves state-of-the-art performance on dense captioning \cite{johnson2016densecap}. The standard evaluation metric for dense captioning combines region proposal and caption generation metrics. Because we aim to generate captions for \emph{any} given region provided by a user (and not just model-proposed ones), we instead evaluate generation capacity \emph{given} a region. Specifically, we conduct a pairwise human evaluation comparing the generations of GRiT on its proposed bounding boxes vs. our model's generations on the same GRiT-proposed bounding boxes. 5 authors of this work evaluated 150 randomly-sampled captioned regions from test set examples in a head-to-head setting. Annotators could select ``A", ``B", or ``Tie": GRiT and our region captioner were randomly assigned A or B placement in a blinded fashion. Overall: while both performed well, our region captioner was preferred to GRiT on average. In 46\% (347/750) of cases, annotators reported a tie, in 34\% (248/750) of cases, annotators reported ours was better, and in 19\% (145/750) of cases, GRiT was rated as better.

Given the (image, region) $\rightarrow$ description model, we next sample candidate regions of interest; in \S~\ref{sec:sec_with_generation_process}, we condition on these regions for the generation of commonsense knowledge. We use the ViT-H Cascade Mask R-CNN \cite{li2022exploring} trained on LVIS \cite{gupta2019lvis} for an initial proposal set. The detector outputs up to 300 candidate objects per image, many of which overlap or cover background objects that are not the focus of the scene. For each image's list of candidate objects, we heuristically downsample to a set of ``most interesting" regions by: 1) selecting the at-most $k=4$ largest/most central people; 2) keeping the most central/large objects; 3) over-sampling rarer objects according to prior frequency of detection in the LVIS vocabulary; 4) limiting the number of objects of a single type per-image; and 5) downsampling overlapping region proposals to encourage broader coverage of the pixel area of the image.

\section{Human Annotation Details}
All human evaluations are performed using the Amazon Mechanical Turk (MTurk) platform. 218 workers from English native speaking countries, at least 5,000 HITs, and acceptance rate $\ge$ 50, are selected based on their passing performance on a paid qualification HIT. The workers are paid with an average rate of \$15/hour. An IRB exemption was obtained for the institution's internal institutional review and ethics board, and we did not collect any denanonymizing information nor do we publish with our dataset sensitive information such as MTurk
IDs.  

\begin{figure*}[htb!]
\centering
\includegraphics[trim=2cm 1cm 0cm 0cm, width=0.99\linewidth]{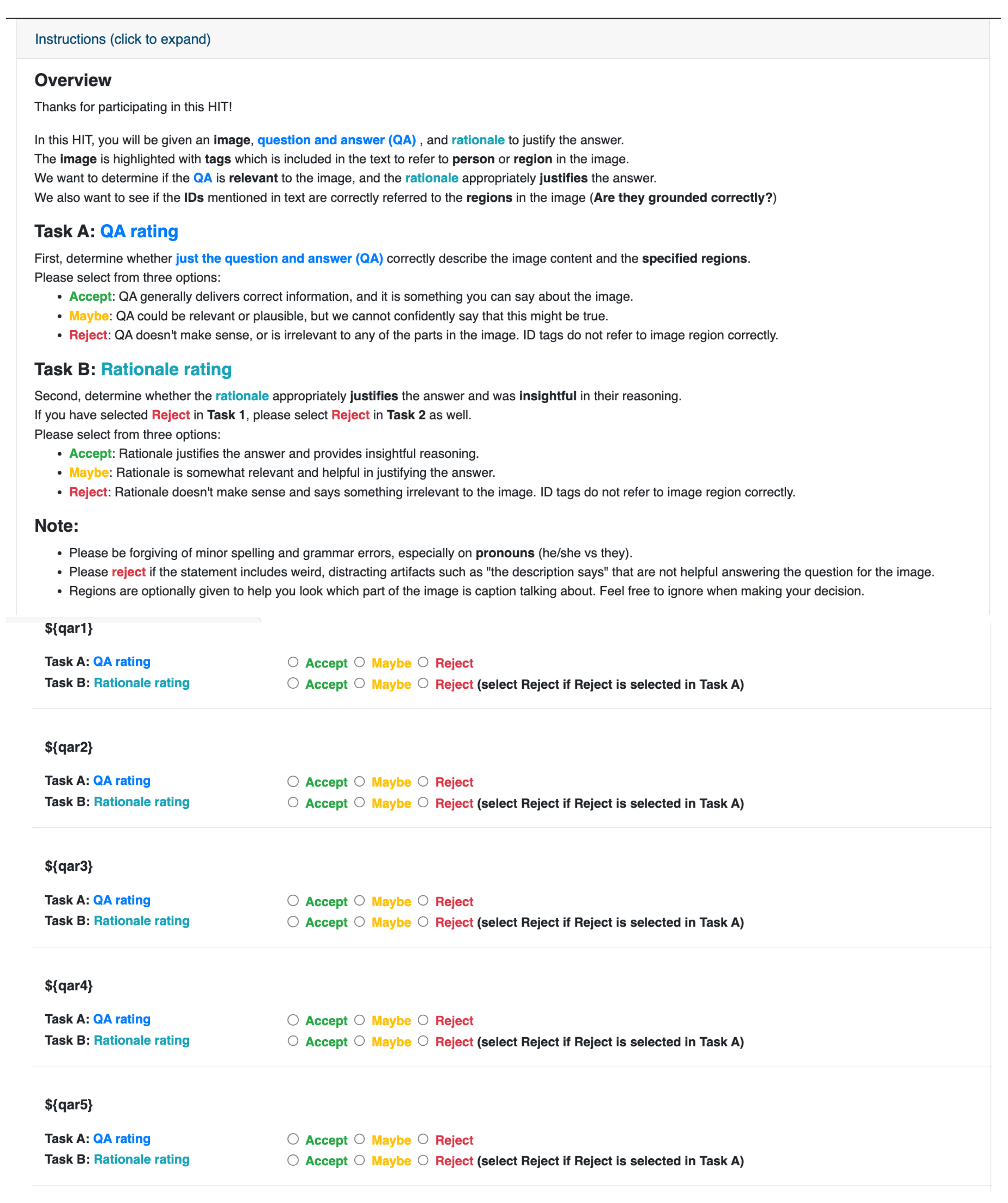}
\caption{Template for acceptability annotation to train the critic model.
}
\label{fig:critic_template}
\end{figure*}

We collect acceptability labels for critic training using the template in Figure ~\ref{fig:critic_template}. For each image and its set of annotated question, answer, rationales (QARs), we run deduplication by clustering the QAR's using hierarchical clustering\footnote{We use the scipy library \url{https://docs.scipy.org/doc/scipy/reference/cluster.hierarchy}.} with their semantic similarity measured by the SentBert \texttt{paraphrase-MiniLM-L6-v2} model ~\cite{reimers-2019-sentence-bert}. We select five question, answer, and rationale triples by getting the roots of the fiver clusters and considering them as the annotation candidates for each image. Using 4,800 images and 24K QAR's, we run the annotation pipeline following Section 2.3 and acquire the acceptability labels for the critic.

\begin{figure*}[htb!]
\centering
\includegraphics[trim=2cm 1cm 0cm 0cm, width=0.99\linewidth]{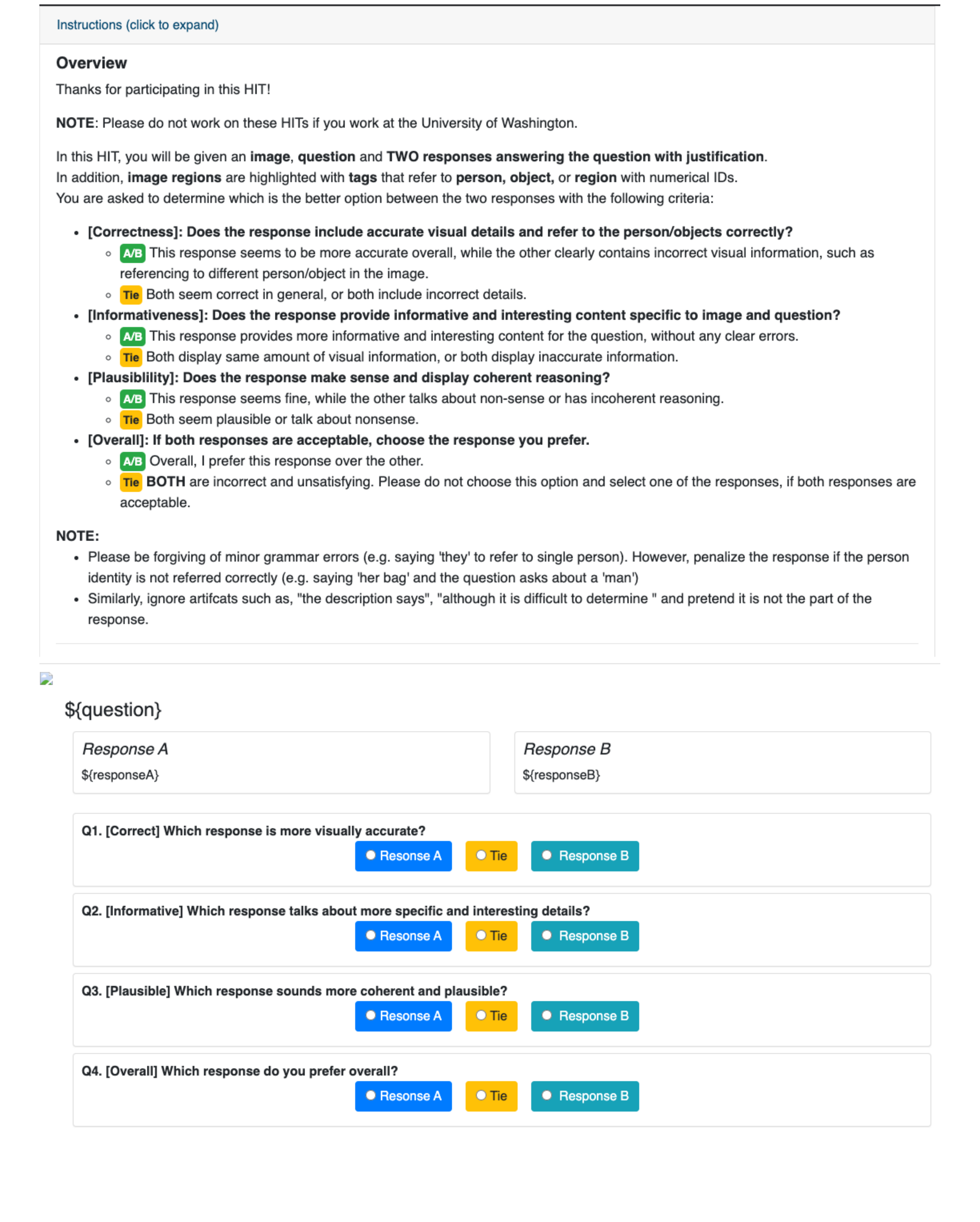}
\caption{Template for pairwise human evaluation.
}
\label{fig:pairwise}
\end{figure*}

Figure ~\ref{fig:pairwise} shows the template to conduct the pairwise human evaluation comparing ours to chatgpt responses with VCR questions and images ~\cite{zellers2019recognition}. To reduce the label selection bias, we randomize the order of two responses. 300 (image, QAR) pairs are selected for evaluation where there is no overlap among the images. Three annotators are asked to make a selection, and instances that did not receive at least two votes are not considered in each evaluation criteria, which we found to be 6\% on average.  

\section{Additional Qualitative Examples}
\begin{figure*}[htb!]
\centering
\includegraphics[trim=2cm 1cm 0cm 0cm, width=0.99\linewidth]{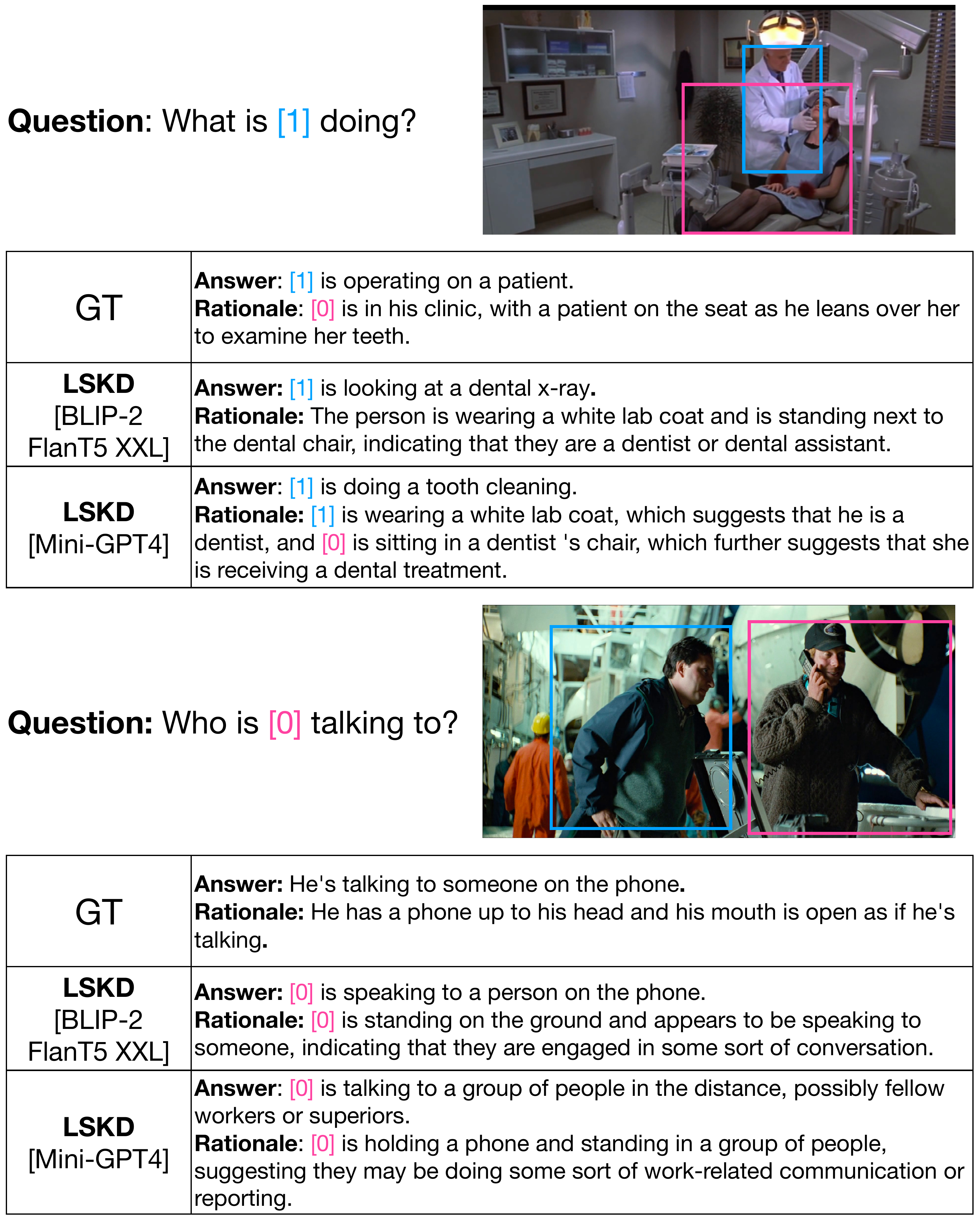}
\caption{Qualitative Examples generated with different models trained with \method.}
\label{fig:more_qual_examples}
\end{figure*}

In Figure ~\ref{fig:more_qual_examples}, we present qualitative results of BLIP-2 FlanT5$_\mathrm{XXL}$ and Mini-GP4 models trained with \method, for answering VCR questions \cite{zellers2019recognition}. The results demonstrate that both models are capable of accurately identifying the relevant person performing the action. For instance, in the first example, the models correctly identify [1] as a dentist due to the presence of a lab coat. Similarly, in the second example, they recognize that [0] is the individual talking on the phone. Notably, the Mini-GPT4 model, which employs the more powerful language model Vicuna \cite{vicuna2023}, produces more precise answers. For instance, it mentions specific actions like tooth cleaning and identifies [0] as seated in the dentist's chair. Additionally, it hypothesizes that [0] might be engaged in conversation with other workers or superiors based on the observation of holding a phone. This observation suggests that \method benefits from employing a language model with enhanced capabilities as indicated by the human evaluation results in the main paper.

\begin{figure*}[htb!]
\centering
\includegraphics[trim=2cm 1cm 0cm 0cm, width=0.99\linewidth]{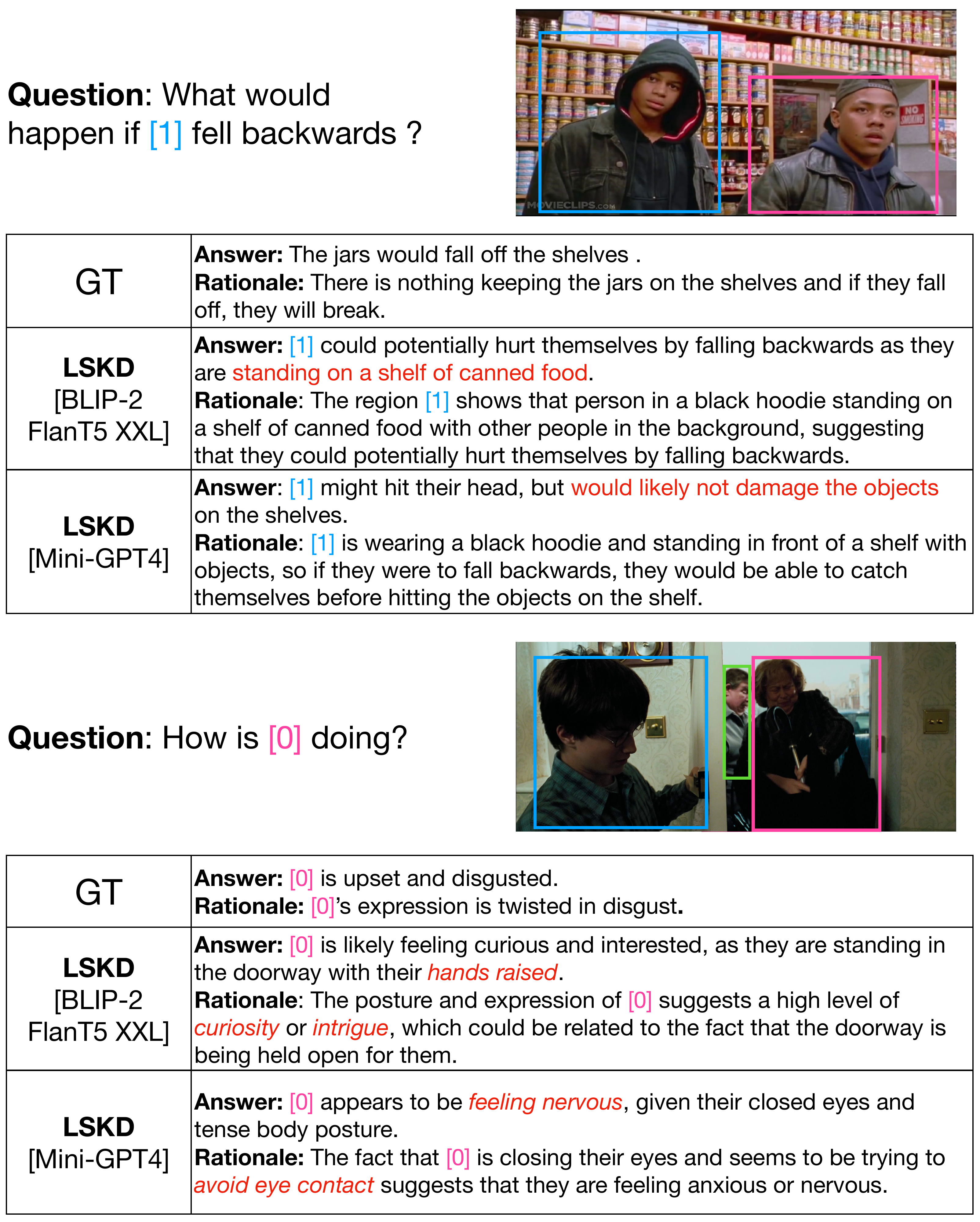}
\caption{Examples with minor errors in spatial reasoning and expression comprehension.
}
\vspace{0.5cm}
\label{fig:failure}
\end{figure*}

We also show failure cases in Figure ~\ref{fig:failure}. We observe that the models are capable of correctly identifying the individuals, such as recognizing [1] as the person wearing a black hoodie and [0] as the individual with closed eyes standing in the doorway. However, they 1) demonstrate a lack of spatial reasoning. For instance, the T5 model hallucinates that the boy is "standing on a shelf of canned food," while Mini-GPT4 hypothesizes that he would "not damage the objects" if he were to fall over, despite the close proximity of the objects in the first example. Additionally, in the second example, the models exhibit a 2) deficiency in fine-grained understanding of people's expressions. Although [0] displays a disgusted facial expression, the T5 model incorrectly interprets it as curiosity and interest, while Mini-GPT4 predicts that she is feeling nervous. These observations indicate that while the models are able to correctly identify the relevant regions, they still lack the capability for nuanced and intricate understanding that necessitates more sophisticated reasoning of visual content.

\section{Error Bars}
We report error bars for the BLIP-2 \cite{li2023blip} trained with \method{} in Table 2 of the main paper. We run three experiments with different random seeds and show the results in Table ~\ref{tab:error_bar}. 
Note all other methods are evaluated with off-the-shelf zero-shot models, hence we only report error bars just for our method. 
\begin{table*}[htb!]
\normalsize
\centering
\scalebox{0.9}{
\begin{tabular}{l | c c c | c | c}
\toprule

\textbf{} & \multicolumn{3}{c}{\textbf{VCR}} & \multicolumn{1}{c}{\textbf{Sherlock}} & 
{\textbf{VisualCOMET}} \\

\textbf{} & Q $\rightarrow$ A & QA $\rightarrow$ R & Q $\rightarrow$ AR & Comparison & Acc@50 \\ 
\midrule
BLIP-2 ViT-G + \method & 58.8 $\pm$ 0.12 & 56.3 $\pm$ 0.07 & 33.2 $\pm 0.09$ & 30.1 $\pm$ 0.09 & 40.0 $\pm$ 0.11 \\ 

\bottomrule
\end{tabular}
}
\caption {
    Error bars of \method{} on zero-shot localized visual reasoning tasks (last row of Table 2).
} 
\label{tab:error_bar}
\vspace{0.5mm}
\end {table*}

\section{Limitations}
One limitation is the recognition bottleneck of the verbalizers, in which off-the -shelf vision language models may encounter errors in  object detection or action recognition. With a combination of verbalization output, the LLM largely ignores irrelevant and incoherent content in the image, but is still prone to generating erroneous data. We made efforts to mitigate the issue by training a supervised critic model on a subset of data to filter out erroneous cases. However, it should be noted that the critic model cannot guarantee the exclusion of all irrelevant instances. Despite these limitations, the trained \method{} models exhibit notable improvements and demonstrate impressive generation capabilities when applied to localized visual reasoning tasks.

Another limitation is the coverage of questions in the data. As shown in Table \ref{tab:question_types}, the dataset encompasses various question types; however, there may still exist certain question categories that are underrepresented or not adequately captured (\eg object counts, potential challenges, other inferences). This limitation could potentially affect the generalizability of the models trained on the dataset to specific question types that are underrepresented or absent.


\end{document}